\documentclass{article}

\PassOptionsToPackage{numbers,compress}{natbib}

 \usepackage[main, final]{neurips_2026}

\usepackage[utf8]{inputenc} %
\usepackage[T1]{fontenc}    %
\usepackage{hyperref}       %
\usepackage{url}            %
\usepackage{booktabs}       %
\usepackage{amsfonts}       %
\usepackage{nicefrac}       %
\usepackage{microtype}      %
\usepackage{xcolor}         %
\usepackage{graphicx}
\usepackage{caption}
\usepackage{amsfonts, amsmath}
\usepackage{color,soul}
\usepackage{enumitem}
\usepackage{cleveref} 
\usepackage{bbm}
\usepackage[table]{xcolor}
\usepackage{array}
\usepackage{booktabs}
\usepackage{multirow}
\usepackage{graphicx}
\usepackage{arydshln}
\usepackage{wrapfig}

\newcommand{\rgb}{\mathrm{rgb}}
\newcommand{\thr}{\mathrm{thr}}
\newcommand{\tcon}{\mathrm{c}}
\newcommand{\thf}{\mathrm{h}}

\newcommand{\AvgPool}{\operatorname{AvgPool}}
\newcommand{\HP}{\operatorname{HP}}
\newcommand{\sg}{\operatorname{sg}}

\newcommand{\bestcell}[1]{\cellcolor{red!25}#1}
\newcommand{\secondcell}[1]{\cellcolor{orange!25}#1}
\newcommand{\thirdcell}[1]{\cellcolor{yellow!35}#1}

\definecolor{lightpurple}{RGB}{150,120,210}

\newcolumntype{C}[1]{>{\centering\arraybackslash}m{#1}}

\newlength{\darklabelw}
\newlength{\darkmodew}
\newlength{\darkrowh}

\newlength{\darkmidh}

\title{DarkVGGT: Seeing Through Darkness \\ Using Thermal Geometry}

\author{%
Minseong Kweon$^{1}$ \quad
Wenyuan Zhao$^{2}$ \quad
Nuo Chen$^{2}$ \quad
Lulin Liu$^{1}$ \\[0.5ex]
\textbf{Huiwen Han}$^{3}$ \quad
\textbf{Zihao Zhu}$^{2}$ \quad
\textbf{Srinivas Shakkottai}$^{2}$ \quad
\textbf{Chao Tian}$^{2}$ \quad
\textbf{Zhiwen Fan}$^{2}$ \\[1ex]
$^{1}$University of Minnesota \qquad
$^{2}$Texas A\&M University \qquad
$^{3}$Stanford University
}

\begin{document}

\maketitle

\vspace{-2.em}
\begin{center}
    \small
    \textbf{\textcolor{purple!80}{Project Page: }}%
    {\hypersetup{urlcolor=black}\url{https://darkvggt.github.io}}
\end{center}

\begin{center}
    \includegraphics[width=0.99\linewidth]{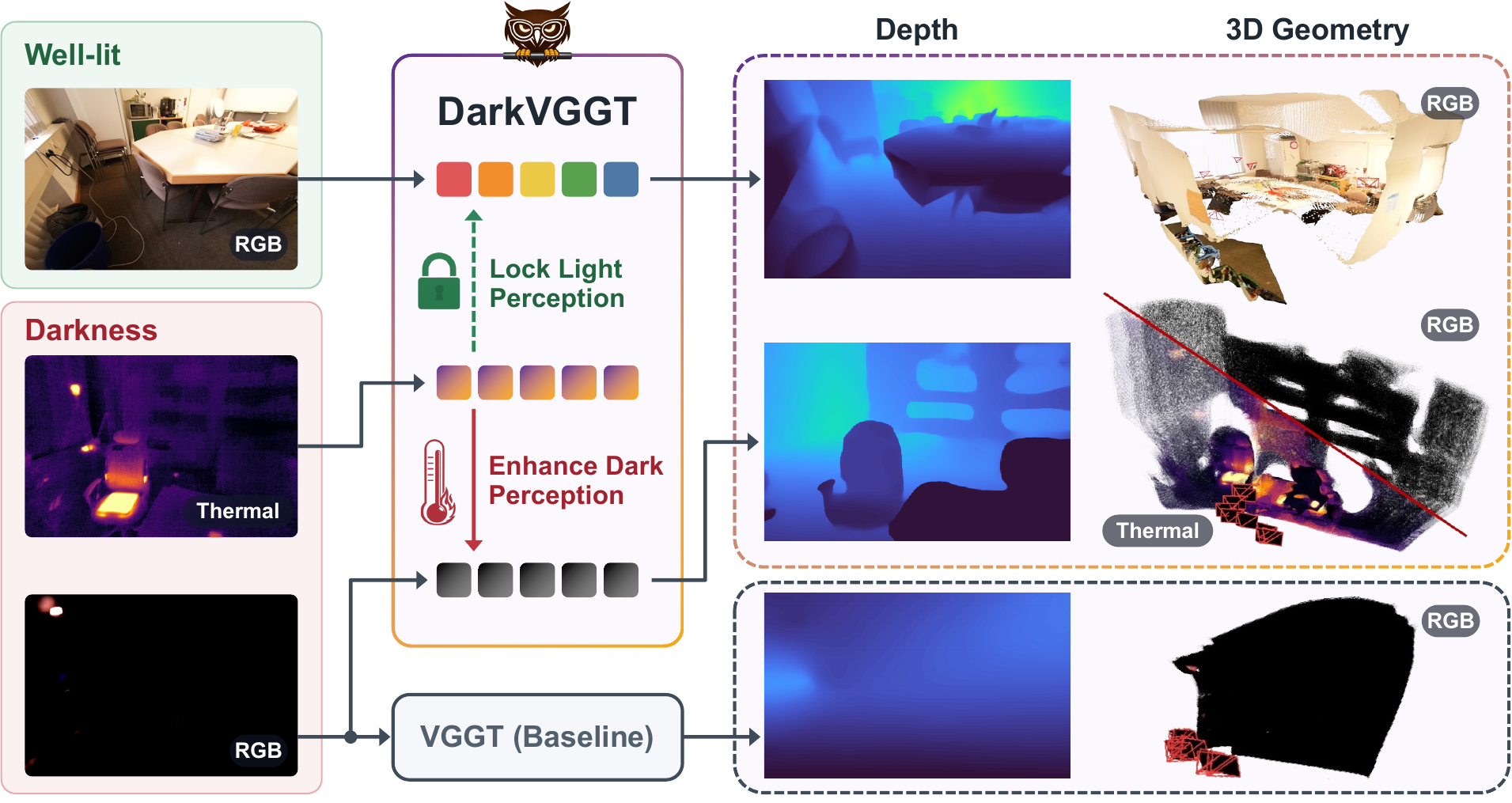}
    \captionof{figure}{DarkVGGT achieves robust low-light 3D geometry estimation through effective thermal-to-RGB fusion. It performs reliably in dark environments where existing feed-forward foundation models often fail, while largely preserving well-lit performance.}
    \label{fig:teaser}
\end{center}

\begin{abstract}
Recent feed-forward 3D reconstruction methods have demonstrated strong performance and flexibility in efficient end-to-end scene geometry estimation from image streams.
However, their reliance on visible-light appearance makes them vulnerable in dark and low-visibility environments, where RGB cues are severely degraded and geometric evidence becomes ambiguous.
To address this challenge, we propose DarkVGGT, a structured RGB-T adaptation of pretrained VGGT for robust 3D estimation in low-light scenes.
DarkVGGT introduces two complementary modules. First, complementary thermal feature factorization processes thermal tokens through content-preserving and high-frequency residual pathways controlled by learned complementary allocation weights.
Second, geometry-shared thermal routing aligns RGB and thermal representations in a shared latent space and selectively injects a gated feature correction into the RGB stream.
Together, these components enable accurate thermal-informed geometry estimation under degraded RGB conditions while largely preserving performance in well-lit environments.
Experiments on low-visibility RGB-T benchmarks demonstrate consistent improvements in both depth and camera pose estimation over existing feed-forward geometry baselines.

\end{abstract}

\section{Introduction}
\label{sec:introduction}

Feed-forward visual geometry models have emerged as a promising alternative to conventional SfM~\citep{schonberger2016structure} and MVS~\citep{schonberger2016pixelwise} pipelines, directly predicting camera parameters and dense 3D structure from image streams with efficient inference.
DUSt3R~\citep{wang2024dust3r} demonstrates this paradigm through dense point-map regression, and subsequent work has extended it to matching~\citep{leroy2024grounding,duisterhof2025mast3r}, dynamic scenes~\citep{zhang2024monst3r,han2025d,fang2026more}, and scalable multi-view aggregation~\citep{wang20253d,cabon2025must3r,yang2025fast3r}.
VGGT~\citep{wang2025vggt} further unifies this direction by jointly predicting cameras, depth, point maps, and point tracks within a single feed-forward framework.
These advances have broadened the applicability of feed-forward geometry models to robot navigation and manipulation~\citep{yang2026robo3r}, autonomous driving~\citep{lu2024drivingrecon,cho2025vr}, and real-time SLAM~\citep{selvaratnam20253d,murai2025mast3r,maggio2025vggt}.
However, most existing models remain centered on well-lit RGB inputs, while reliable autonomous systems must also operate under challenging illumination, including nighttime roads and dark indoor scenes~\citep{arnold2019survey,ghari2024pedestrian}.
Under low visibility, degraded RGB appearance and correspondence cues limit the reliability of feed-forward geometry models pretrained on well-lit images~\citep{he2023darkfeat}.

Recent methods have improved feed-forward geometry prediction in low-visibility scenes through clean-to-noisy distillation of image matching capabilities~\citep{guo2026dark3r}, event-camera-based feed-forward geometry modeling~\citep{ren2026eventvggt,wu2025eag3r}, and thermal sensing with infrared cameras~\citep{skorokhodov2026sear}.
These methods show that low-light adaptation or additional sensing can recover geometric cues when RGB images become unreliable.
Among these directions, thermal imaging is especially appealing for dark-scene geometry estimation because long-wave infrared radiation is largely insensitive to visible illumination and complements RGB cues when texture and color evidence degrade.

However, robust low-light performance alone is not sufficient for a reliable autonomous perception system.
Such a system must improve 3D estimation under degraded illumination while preserving the geometric capability learned from standard well-lit RGB data.
This requirement remains difficult because existing low-visibility adaptations may disturb the pretrained RGB geometry prior and reduce well-lit 3D perception, while multisensor inputs can introduce sparse or noisy modality-specific cues~\citep{zhang2026multimodal}.
RGB-T fusion therefore needs to use thermal information as selective corrective evidence rather than as an unconstrained additional appearance stream.
This is challenging for two reasons.
First, foundation models such as VGGT acquire strong geometric priors from large-scale RGB pretraining, while paired RGB-T data with 3D supervision remains limited~\citep{maheshwari2026anythermal,brenner2023rgb}.
Naive RGB-T mixing can therefore erode the pretrained well-lit representation and create a ``daylight tax'', where low-light gains come at the cost of degraded well-lit performance.
Second, thermal appearance can vary independently of visible appearance and depth structure because of temperature, material properties, and background radiation~\citep{usamentiaga2014infrared}.
Thus, RGB-T fusion should selectively modulate thermal features and route task-relevant corrections into the RGB geometry stream.

We therefore propose \textbf{DarkVGGT}, a feed-forward RGB-T geometry framework for robust 3D perception in dark and low-visibility scenes with low daylight performance degradation.
DarkVGGT treats the thermal stream as a selective source of corrective evidence, rather than an unconstrained appearance channel.
Within the proposed \emph{Alternating Attention blocks}, it combines complementary thermal feature factorization with geometry-shared thermal routing: thermal features are processed through content-preserving and high-frequency residual pathways, while gated corrections in a shared latent space are routed to the RGB stream.
This geometric adaptive fusion design uses thermal observations when RGB cues are unreliable, while preserving the pretrained RGB geometry prior and maintaining strong performance in well-lit scenes.
Our contributions are summarized below:
\vspace{-0.5em}

\begin{itemize}[leftmargin=2em]
\item We formulate dark-scene RGB-T geometry estimation as a feed-forward multisensor fusion problem and propose \textbf{DarkVGGT}, a structured RGB-T adaptation of pretrained VGGT that retains the RGB geometry as the primary estimator while using thermal observations to provide gated corrections under low-light conditions.

\item We introduce an RGB-T fusion strategy that combines complementary thermal feature factorization with geometry-shared thermal routing, providing complementary pathways of a content-preserving branch and a high-frequency residual branch, and selectively injecting corrections from a shared feature subspace to improve dark-scene geometry perception.

\item DarkVGGT achieves strong depth and camera pose estimation performance across diverse low-light datasets, including ViViD++, STheReO, and Dark3R, while preserving geometric estimation performance on standard well-lit benchmarks such as ETH3D and ScanNet++, demonstrating the effectiveness of the proposed thermal-to-RGB fusion strategy.
\end{itemize}

\section{Related Work}
\label{sec:relate-work}

\paragraph{Feed-forward 3D reconstruction.} 
DUSt3R~\citep{wang2024dust3r} marked a significant advance in 3D reconstruction by directly regressing dense point maps from unconstrained pairwise images in an end-to-end framework.
Subsequent works have evolved in several directions, including improving pairwise matching~\citep{leroy2024grounding,duisterhof2025mast3r}, handling dynamic scenes~\citep{zhang2024monst3r}, and accelerating inference by avoiding explicit global correction~\citep{wang20253d,cabon2025must3r}.
VGGT~\citep{wang2025vggt} pushes this line further by jointly predicting cameras, depth, point maps, and 3D tracks in a unified Alternating Attention (AA) transformer, making it a natural foundation model for geometric reasoning.
Beyond RGB-only reconstruction, recent works have begun to explore multimodal or degraded-visibility feed-forward geometry, including low-light structure-from-motion~\citep{guo2026dark3r} and RGB-thermal adaptation of visual geometry transformers~\citep{skorokhodov2026sear}.
Most closely related to our setting, concurrent SEAR~\citep{skorokhodov2026sear} shows that a pretrained VGGT can be adapted to RGB-T inputs using LoRA-based~\citep{hu2022lora} fine-tuning, thermal camera tokens, and tailored batching strategies.
DarkVGGT extends this baseline with complementary thermal feature factorization and geometry-shared thermal routing, selectively injecting thermal feature corrections into the RGB stream.

\vspace{-4mm}
\paragraph{Low-light 3D vision with thermal images.} 
Thermal imaging has long been recognized as complementary to RGB under low illumination, smoke, fog, and nighttime conditions. 
Prior work has explored thermal cues for low-light perception in tasks such as depth estimation, RGB-T sensor fusion, SLAM, and neural scene representation learning~\citep{shin2023deep,shin2021self,chen2024thermal3dgs,kweon2025mrgs,lu2024thermalgaussian}.
A parallel line studies RGB-T or multispectral depth fusion, including cross-spectrum depth estimation from dual-modality cameras~\cite{guo2022cross}, depth estimation from misaligned RGB-T image pairs~\cite{kwon2024misaligned}, confidence-aware RGB-to-thermal distillation~\cite{zuo2025monother}, and the recent align-and-fuse strategy that couples geometry-guided representation correction with an attachable fusion module~\cite{shin2025bridging}.
Recent work on 3D scene modeling and novel-view synthesis jointly models RGB and thermal observations or introduces temperature-consistency and physics-aware priors~\cite{hassan2024thermonerf,lin2024thermalnerf,lu2024thermalgaussian}. 
Dual-band thermal videography~\citep{narayanan2026dual} combines spectral and temporal cues to separate time-varying emission and reflection near ambient conditions. DarkVGGT focuses on learning geometry-oriented feature modulation from normalized single-band thermal inputs.
For dense 3D point reconstruction, SEAR~\citep{skorokhodov2026sear} proposes predicting 3D geometry by fine-tuning VGGT with LoRA on unpaired RGB-T streams. However, paired RGB-T images remain underexplored for dark-scene geometry estimation. DarkVGGT bridges this gap through an effective thermal-to-RGB fusion method.

\section{Method}
\label{sec:method}

\subsection{Overview of DarkVGGT}

Feed-forward visual geometry models directly predict camera parameters, depth, point maps, and tracks from image observations, reducing reliance on conventional SfM/MVS pipelines.
In this work, we leverage VGGT~\citep{wang2025vggt} as the base geometry predictor.
Given a sequence $\{I_s\}_{s=1}^{S}$ of RGB image frames, where $I_s\in\mathbb{R}^{3\times H\times W}$, VGGT first patchifies each image and embeds it into a set of $P$ tokens $\mathbf{x}_s\in \mathbb{R}^{P\times C}$ using DINOv2~\citep{oquab2023dinov2}.
Each frame is augmented with a camera token $\mathbf{c}_s\in \mathbb{R}^{1\times C}$ and four register tokens $\mathbf{r}_s\in\mathbb{R}^{4\times C}$.
The resulting token sequence $\mathbf{t}_s := [\mathbf{c}_s;\mathbf{r}_s;\mathbf{x}_s]$ is processed by frame-wise and global Alternating Attention (AA) blocks, which model both per-view structure and cross-view correspondence.
We denote the RGB patch tokens at the $\ell$-th AA block as $\mathbf{x}^{\rgb,(\ell)} \in \mathbb{R}^{BS \times P \times C}$, where $B$ is the batch size and $C$ is the token dimension.
Task-specific prediction heads, including DPT-style dense prediction heads~\citep{ranftl2021vision}, then recover camera parameters and dense geometry from the shared representation.

Building on this, DarkVGGT introduces RGB-T thermal guidance for robust low-visibility geometry estimation.
Given paired RGB and thermal frames, denoted by $\{(I_s^{\rgb}, I_s^{\thr})\}_{s=1}^{S}$, DarkVGGT predicts camera parameters (e.g., the predicted pose $\hat{P}$) and depth maps $\hat{D}$ on the RGB stream while leveraging thermal observations as an auxiliary modality in dark scenes. The framework builds on a pretrained VGGT backbone~\citep{wang2025vggt} with a frozen DINOv2 encoder~\citep{oquab2023dinov2}. At the $\ell$-th AA block, the patched RGB and thermal tokens are denoted by $\mathbf{x}^{\rgb,(\ell)}, \mathbf{x}^{\thr,(\ell)} \in \mathbb{R}^{BS \times P \times C}$, respectively. For RGB-T adaptation, we use shared LoRA adapters~\citep{hu2022lora} on the AA blocks with modality-specific camera tokens. Details are provided in Appendix~\ref{sec:supp-method}.

\Cref{fig:main_framework} provides an overview of \textbf{DarkVGGT}.
On top of the VGGT-based backbone, DarkVGGT introduces two additional modules:
1) Complementary Thermal Feature Factorization (CTFF) and
2) Geometry-Shared Thermal Routing (GSTR).

\begin{figure}[t]
    \centering
    \includegraphics[width=\linewidth]{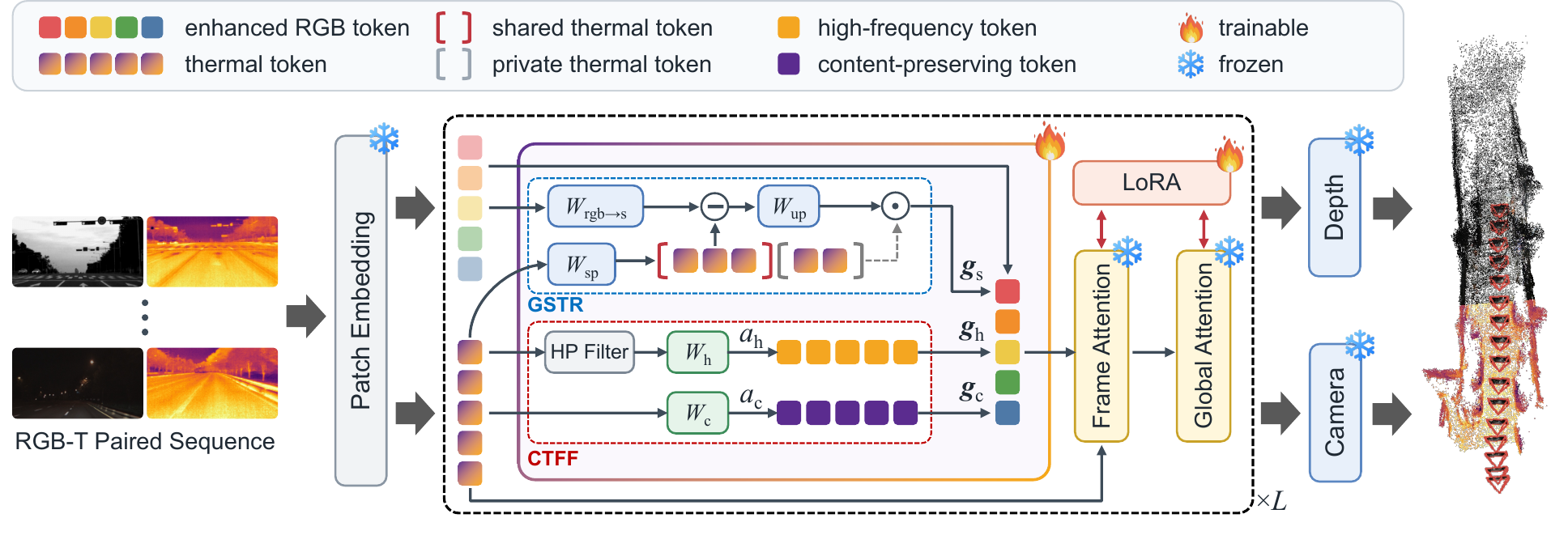}
\caption{Overview of the DarkVGGT framework.
DarkVGGT modulates thermal embeddings through content-preserving and high-frequency residual pathways using a high-pass (HP) filter and projection branches.
Geometry-shared thermal routing (GSTR) learns shared/private thermal representations and injects gated corrections from the shared latent space into the RGB stream.}
    \label{fig:main_framework}
\end{figure}

\subsection{Complementary Thermal Feature Factorization}
\label{sec:method_phys}
\begin{wrapfigure}[13]{r}{0.51\linewidth}
    \centering
    \includegraphics[width=\linewidth]{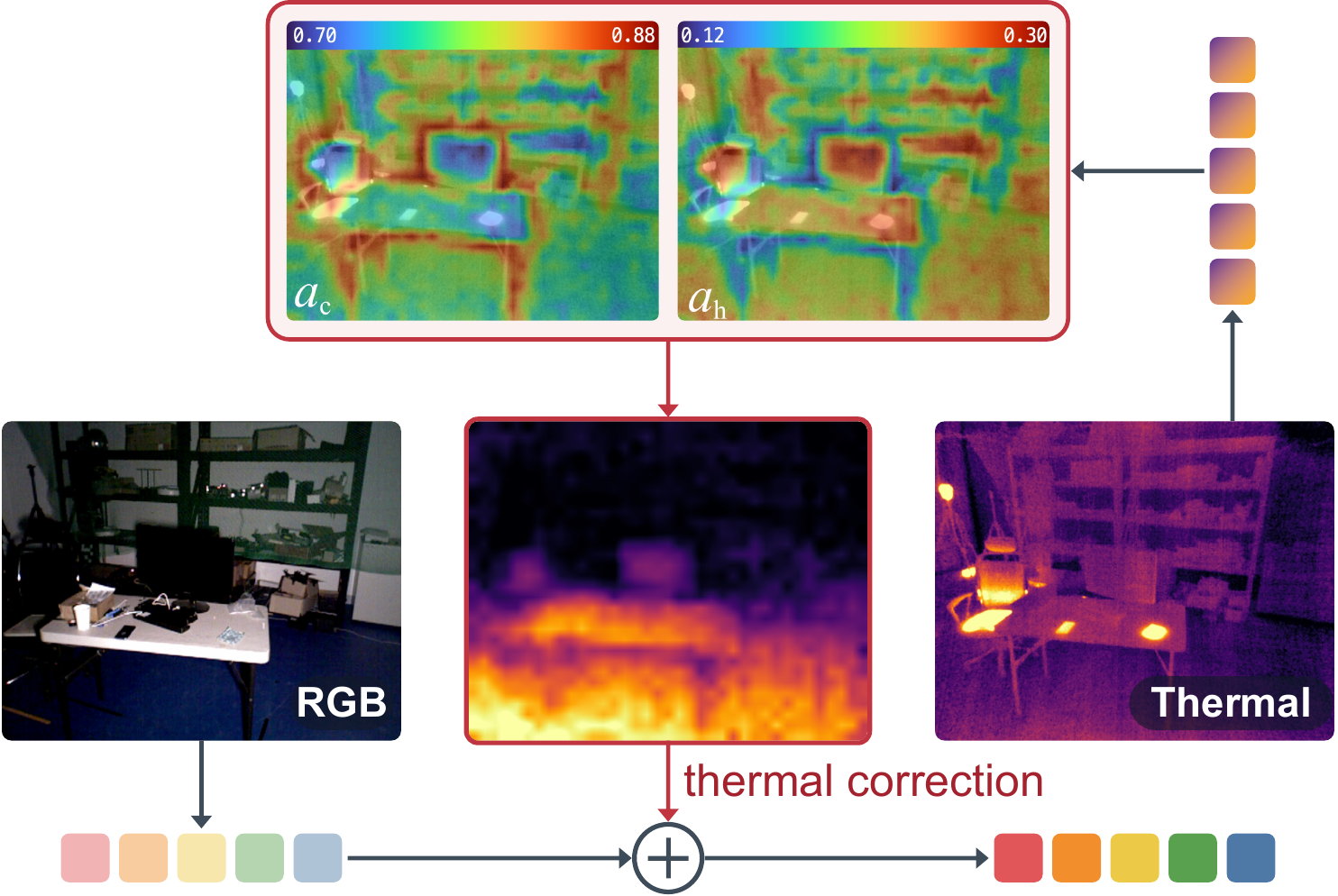}
    \caption{Visualization of complementary thermal feature factorization. Per-patch allocation weights control the content-preserving and high-frequency residual pathways before gated RGB fusion.}
    \label{fig:phys_module}
\end{wrapfigure}

Figure~\ref{fig:phys_module} illustrates Complementary Thermal Feature Factorization (CTFF), which processes thermal tokens through a content-preserving ($\tcon$) pathway and a high-frequency residual ($\thf$) pathway.
Thermal appearance exhibits spatial variations arising from temperature, material properties, and environmental conditions~\citep{usamentiaga2014infrared,Vollmer_2010,narayanan2026dual}.
We process the unfiltered thermal features and their local high-pass residuals through separate learned projections to accommodate these variations.
Complementary allocation weights couple the two pathways, while separate gates regulate their respective contributions to the RGB stream.
The single-band thermal inputs undergo normalization and CLAHE before feature extraction.

\paragraph{Complementary feature modulation.}
At each AA block, we learn a joint RGB-T cross-modal feature $\mathbf{h}^{(\ell)}$ from the joint RGB and thermal patch tokens and use it to predict complementary patch-wise allocation weights:
\begin{equation}
\mathbf{h}^{(\ell)} = \mathrm{FFN}\left([\mathbf{x}^{\rgb,(\ell)}; \mathbf{x}^{\thr,(\ell)}]\right),
\qquad
a_{\mathrm{c}}^{(\ell)} = \sigma\!\left(W_a\mathbf{h}^{(\ell)}\right), 
\quad
a_{\mathrm{h}}^{(\ell)} = 1 - a_{\mathrm{c}}^{(\ell)},
\end{equation}
where $\mathrm{FFN}(\cdot)$ denotes the lightweight feed-forward network used to extract joint RGB-T features. This parameterization determines both allocation weights from a single scalar per patch.
The content-preserving pathway receives unfiltered thermal tokens, while the high-frequency residual pathway emphasizes local spatial variations through a high-pass filter.  
The modulated thermal features are
\begin{equation}
\mathbf{z}^{(\ell)}_{\tcon}
=
a_{\mathrm{c}}^{(\ell)} \odot W_{\tcon}\mathbf{x}^{\thr,(\ell)},
\qquad
\mathbf{z}^{(\ell)}_{\thf}
=
a_{\mathrm{h}}^{(\ell)} \odot W_{\thf}\HP(\mathbf{x}^{\thr,(\ell)}),
\end{equation}
where $W_{\tcon}$ and $W_{\thf}$ are learnable projections for the content-preserving and high-frequency residual pathways, respectively. 
$\HP(\cdot)$ denotes the high-pass filter on the patch grid, implemented as $x-\AvgPool_{3\times3}(x)$. The RGB token is adapted by
\begin{equation}
\tilde{\mathbf{x}}^{\rgb,(\ell)}
= \mathbf{x}^{\rgb,(\ell)}
+ \boldsymbol{g}_{\tcon}^{(\ell)}\odot\mathbf{z}_{\tcon}^{(\ell)}
+ \boldsymbol{g}_{\thf}^{(\ell)}\odot\mathbf{z}_{\thf}^{(\ell)},
\label{eq:graybody-fusion}
\end{equation}
where $\boldsymbol{g}_{k}^{(\ell)}=\sigma(W_{g,k}\mathbf{h}^{(\ell)})$ for $k\in\{\tcon,\thf\}$ denotes the gate. The content-preserving gate is additionally biased by a learned per-patch offset parameterized as a log-variance difference $\Delta u^{(\ell)}=\log\sigma^{2}_{\rgb}-\log\sigma^{2}_{\thr}$ predicted from $\mathbf{h}^{(\ell)}$.
As shown in Figure~\ref{fig:phys_module}, complementary allocation weights and pathway-specific gates jointly modulate the thermal correction added to the RGB tokens.
To encourage complementary specialization, we introduce three auxiliary regularizers.
The sparsity bias $\mathcal{L}_{\mathrm{sparse}}$ penalizes the averaged high-frequency allocation $a_{\mathrm{h}}^{(\ell)}$, the edge-disagreement loss $\mathcal{L}_{\mathrm{edge}}$ supervises this allocation with a soft target derived from thermal--depth edge disagreement, and the orthogonality penalty $\mathcal{L}_{\mathrm{ortho}}$ discourages redundancy between the two factorized embeddings:
\begin{gather}
\mathcal{L}_{\mathrm{sparse}}
=
\mathbb{E}\!\left[a_{\mathrm{h}}^{(\ell)}\right],
\\
\mathcal{L}_{\mathrm{edge}}
=
\mathrm{BCE}\!\left(
a_{\mathrm{h}}^{(\ell)},
a_{\mathrm{h,tar}}^{(\ell)}
\right),
\\
\mathcal{L}_{\mathrm{ortho}}
=
\mathbb{E}\!\left[
\mathrm{Sim}^2\!\left(
\mathbf{z}^{(\ell)}_{\tcon},
\mathbf{z}^{(\ell)}_{\thf}
\right)
\right],
\end{gather}
where $\mathbb{E}[\cdot]$ denotes expectation over batch, frame, and patch locations, $\mathrm{BCE}(\cdot,\cdot)$ denotes binary cross-entropy, and $\mathrm{Sim}(\cdot,\cdot)$ denotes cosine similarity. The soft target $a_{\mathrm{h,tar}}^{(\ell)}=\sigma(\beta(\bar{E}_{\thr}^{(\ell)}-\bar{E}_{D}^{(\ell)}-\tau))$ is derived from thermal--depth edge disagreement, with $\bar{E}_{\thr}^{(\ell)}$ and $\bar{E}_{D}^{(\ell)}$ denoting patch-pooled thermal and ground-truth depth edge magnitudes, $\beta$ controlling target sharpness, and $\tau$ setting the disagreement margin.

\subsection{Geometry-Shared Thermal Routing}
\label{sec:method_gstr}

While thermal cues can compensate for degraded RGB in dark scenes, they contain both geometry-relevant and modality-specific components~\citep{usamentiaga2014infrared}. We introduce a Geometry-Shared Thermal Routing module that encourages cross-modal alignment in a shared latent space via stop-gradient distillation and uses a private branch for thermal reconstruction.

\paragraph{Geometry-shared thermal decomposition.}
Given RGB and thermal patch tokens $\mathbf{x}^{\rgb,(\ell)}$ and $\mathbf{x}^{\thr,(\ell)}$ at an $\ell$-th AA block, we decompose the thermal tokens into geometry-shared and thermal-private components~\citep{wu2021text} and project RGB into the geometry-shared space:
\begin{equation}
[\mathbf{u}^{\mathrm{s},(\ell)};\,\mathbf{u}^{\mathrm{p},(\ell)}]
=
W_{\mathrm{sp}}\,\mathbf{x}^{\thr,(\ell)},
\qquad
\mathbf{v}^{\mathrm{s},(\ell)}
=
W_{\rgb\rightarrow\mathrm{s}}\,\mathbf{x}^{\rgb,(\ell)}.
\end{equation}
Here, $\mathbf{u}^{\mathrm{s},(\ell)}$ and $\mathbf{u}^{\mathrm{p},(\ell)}$ denote thermal features trained for cross-modal alignment and thermal reconstruction, respectively, and $\mathbf{v}^{\mathrm{s},(\ell)}$ denotes the RGB feature projected into the geometry-shared space. We use $\bar{\mathbf{u}}^{\mathrm{s},(\ell)}$, $\bar{\mathbf{u}}^{\mathrm{p},(\ell)}$, and $\bar{\mathbf{v}}^{\mathrm{s},(\ell)}$ to denote their layer-normalized realizations.

\paragraph{Thermal routing supervision.}
With the geometry-shared thermal branch as a stop-gradient teacher, modality-specific projection heads followed by $L_2$ normalization map $\bar{\mathbf{v}}^{\mathrm{s},(\ell)}, \bar{\mathbf{u}}^{\mathrm{s},(\ell)}$ into a common embedding space, yielding $\check{\mathbf{v}}^{\mathrm{s},(\ell)}, \check{\mathbf{u}}^{\mathrm{s},(\ell)}$:
\begin{equation}
\mathcal{L}_{\mathrm{distill}}
=
\mathbb{E}\!\left[
1-
\left\langle
\check{\mathbf{v}}^{\mathrm{s},(\ell)},
\sg\!\left[\check{\mathbf{u}}^{\mathrm{s},(\ell)}\right]
\right\rangle
\right].
\label{eq:l-distill}
\end{equation}
where $\sg[\cdot]$ blocks gradients on the teacher branch (see Appendix~\ref{sec:supp-method} for projection-head dimensions).
We then encourage the private branch to retain thermal input information by reconstructing the thermal tokens with a variance-normalized regression loss:
\begin{equation}
\mathcal{L}_{\mathrm{recon}}
=
\frac{
\mathrm{MSE}\!\left(W_{\mathrm{rec}}\mathbf{u}^{\mathrm{p},(\ell)},\,\sg[\mathbf{x}^{\thr,(\ell)}]\right)
}{
\mathrm{Var}\!\left(\sg[\mathbf{x}^{\thr,(\ell)}]\right)+\epsilon
}.
\end{equation}
where $W_{\mathrm{rec}}$ maps the thermal-private features back to the thermal token space, and $\epsilon$ is a small constant for numerical stability. Finally, to discourage redundancy between the geometry-shared and thermal-private subspaces, we add a small cross-branch decorrelation loss $\mathcal{L}_{\mathrm{decorr}} = \| (\mathbf{u}^{\mathrm{s},(\ell)})^{\top} \mathbf{u}^{\mathrm{p},(\ell)}\|^2_F$.

\paragraph{Gated shared-feature correction.}
The RGB stream is updated using the difference between thermal and RGB features in the shared latent space, while the thermal-private branch supports thermal reconstruction and gate conditioning.
We compute a per-token feature-conditioned gate $\boldsymbol{g}^{(\ell)}_{\mathrm{s}}$ by comparing the RGB geometry-shared and thermal-private representations,
\begin{equation}
\Delta u^{(\ell)}_{\mathrm{sp}}
=
W_{\sigma}^{\rgb}\,\bar{\mathbf{v}}^{\mathrm{s},(\ell)}
-
W_{\sigma}^{\thr}\,\bar{\mathbf{u}}^{\mathrm{p},(\ell)},
\qquad
\boldsymbol{g}^{(\ell)}_{\mathrm{s}}
=
\sigma\!\left(
W_{\mathrm{g}}[\bar{\mathbf{v}}^{\mathrm{s},(\ell)};\bar{\mathbf{u}}^{\mathrm{s},(\ell)}]
+
\Delta u^{(\ell)}_{\mathrm{sp}}
\right),
\end{equation}
where $\Delta u^{(\ell)}_{\mathrm{sp}}$ is a learned offset that modulates the gate using geometry-shared and thermal-private features.
We then route the gated geometry-shared thermal residual to the RGB tokens:
\begin{equation}
\tilde{\mathbf{x}}^{\rgb,(\ell)}
=
\mathbf{x}^{\rgb,(\ell)}
+
\boldsymbol{g}^{(\ell)}_{\mathrm{s}}\odot
W_{\mathrm{up}}\!\left(\sg[\bar{\mathbf{u}}^{\mathrm{s},(\ell)}]-\bar{\mathbf{v}}^{\mathrm{s},(\ell)}\right),
\label{eq:gstr-injection}
\end{equation}
where $W_{\mathrm{up}}$ projects the geometry-shared latent space back to the token space, and $\tilde{\mathbf{x}}^{\rgb,(\ell)}$ denotes the enhanced RGB tokens.
In practice, this module is instantiated only in the final $k$ AA blocks.

\begin{figure*}[t]
    \centering
    \small
    \setlength{\tabcolsep}{0pt}

    \begin{tabular}{@{}c@{}c@{\hspace{0.02\textwidth}}c@{}}
        & \text{Dark3R} & \text{VGGT} \\[0.2em]

        \makebox[0.03\textwidth][l]{%
            \rotatebox[origin=c]{90}{\text{RGB}}%
        }%
        &
        \begin{minipage}[c]{0.475\textwidth}
            \noindent
            \begin{minipage}[c]{0.32\linewidth}
                \centering
                \makebox[\linewidth][c]{%
                    \scalebox{1.04}[1]{%
                        \includegraphics[
                            width=\linewidth
                        ]{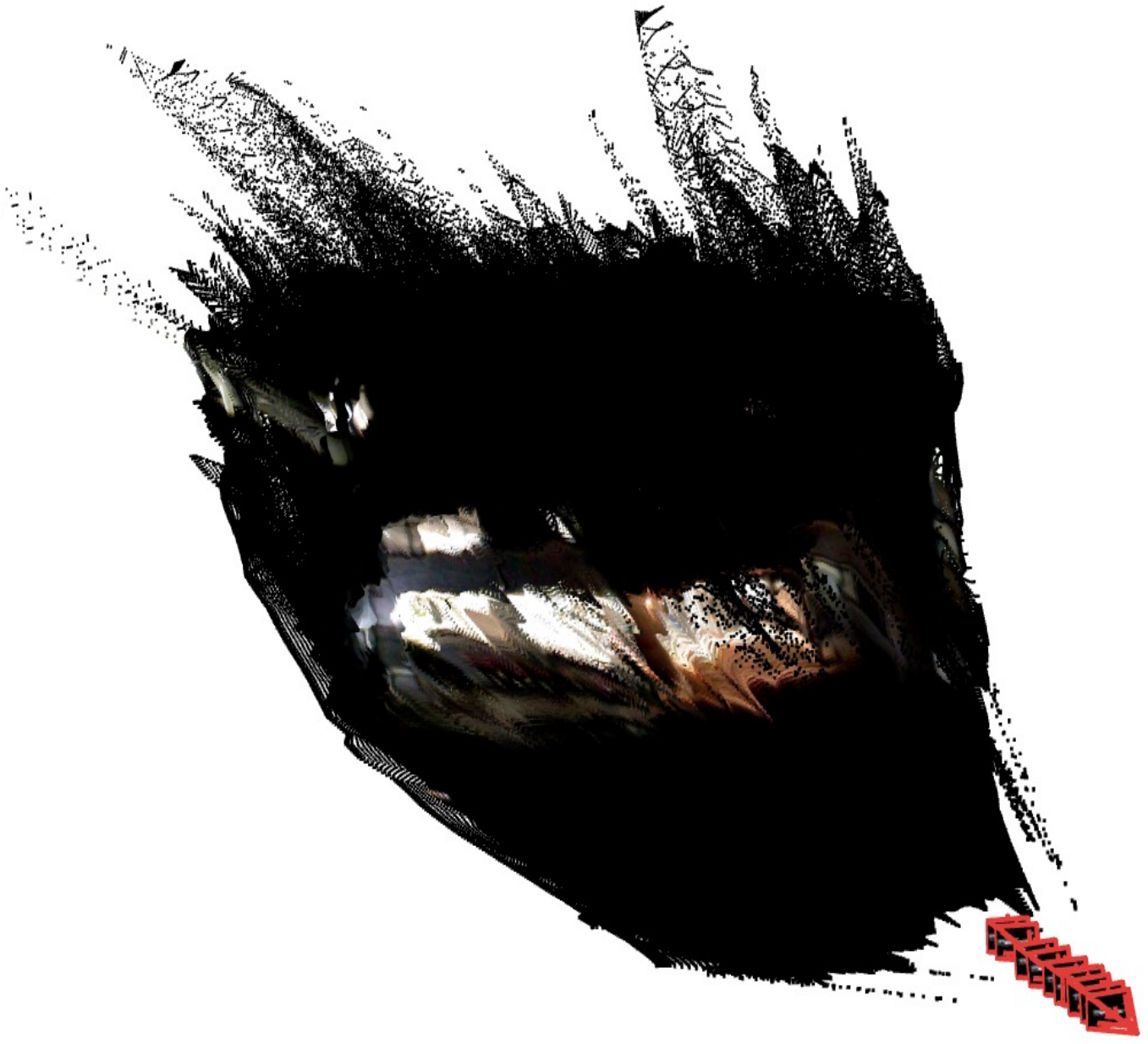}%
                    }%
                }%
            \end{minipage}%
            \hfill
            \begin{minipage}[c]{0.34\linewidth}
                \centering
                \scalebox{1}[1]{%
                    \includegraphics[
                        width=\linewidth
                    ]{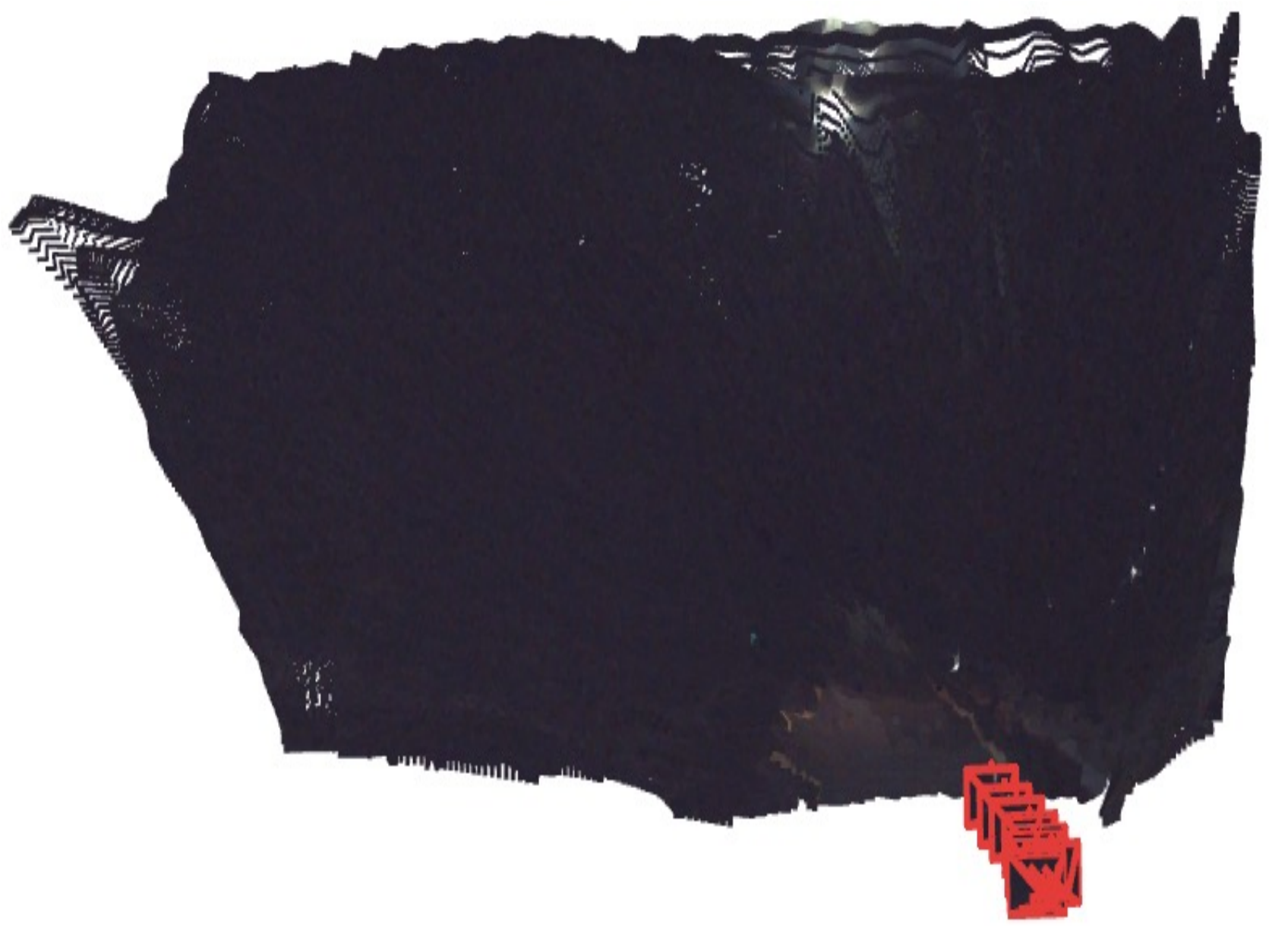}%
                }%
            \end{minipage}%
            \hfill
            \begin{minipage}[c]{0.32\linewidth}
                \centering
                \makebox[\linewidth][c]{%
                    \includegraphics[
                        width=1.04\linewidth,
                        height=0.755\linewidth
                    ]{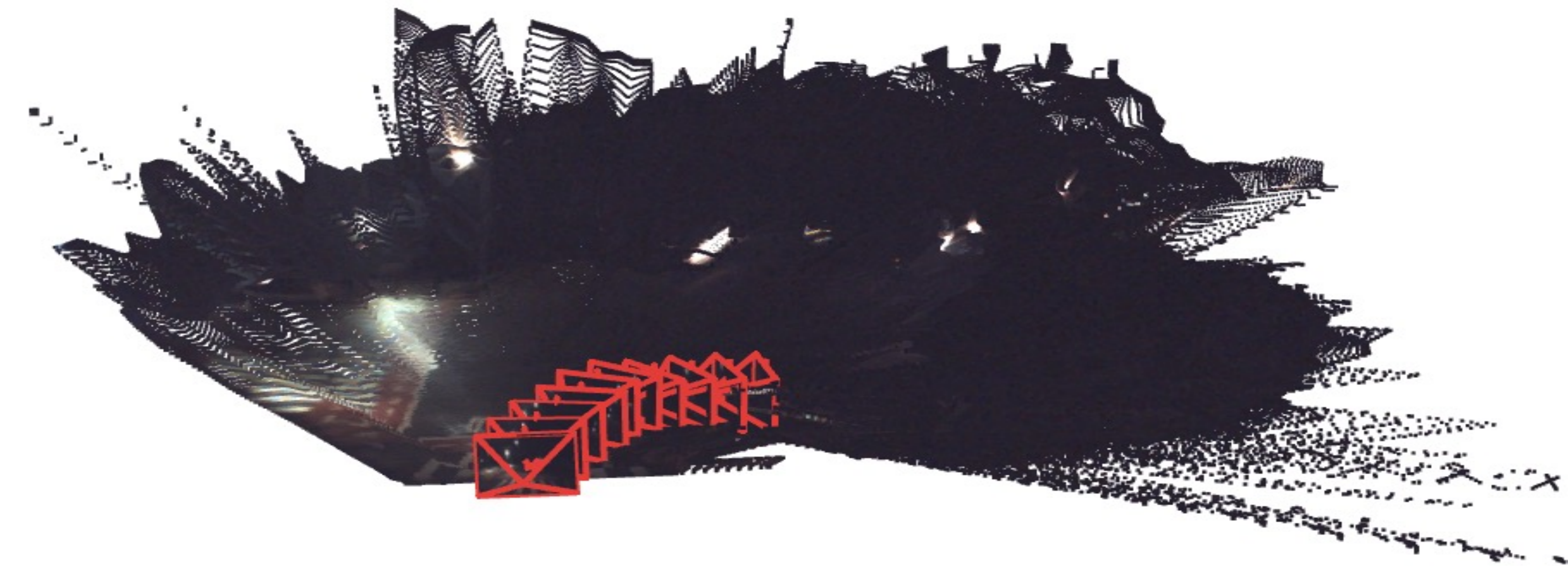}%
                }%
            \end{minipage}%
        \end{minipage}%
        &
        \begin{minipage}[c]{0.475\textwidth}
            \noindent
            \begin{minipage}[c]{0.32\linewidth}
                \centering
                \makebox[\linewidth][c]{%
                    \scalebox{1}[0.90]{%
                        \includegraphics[
                            width=\linewidth
                        ]{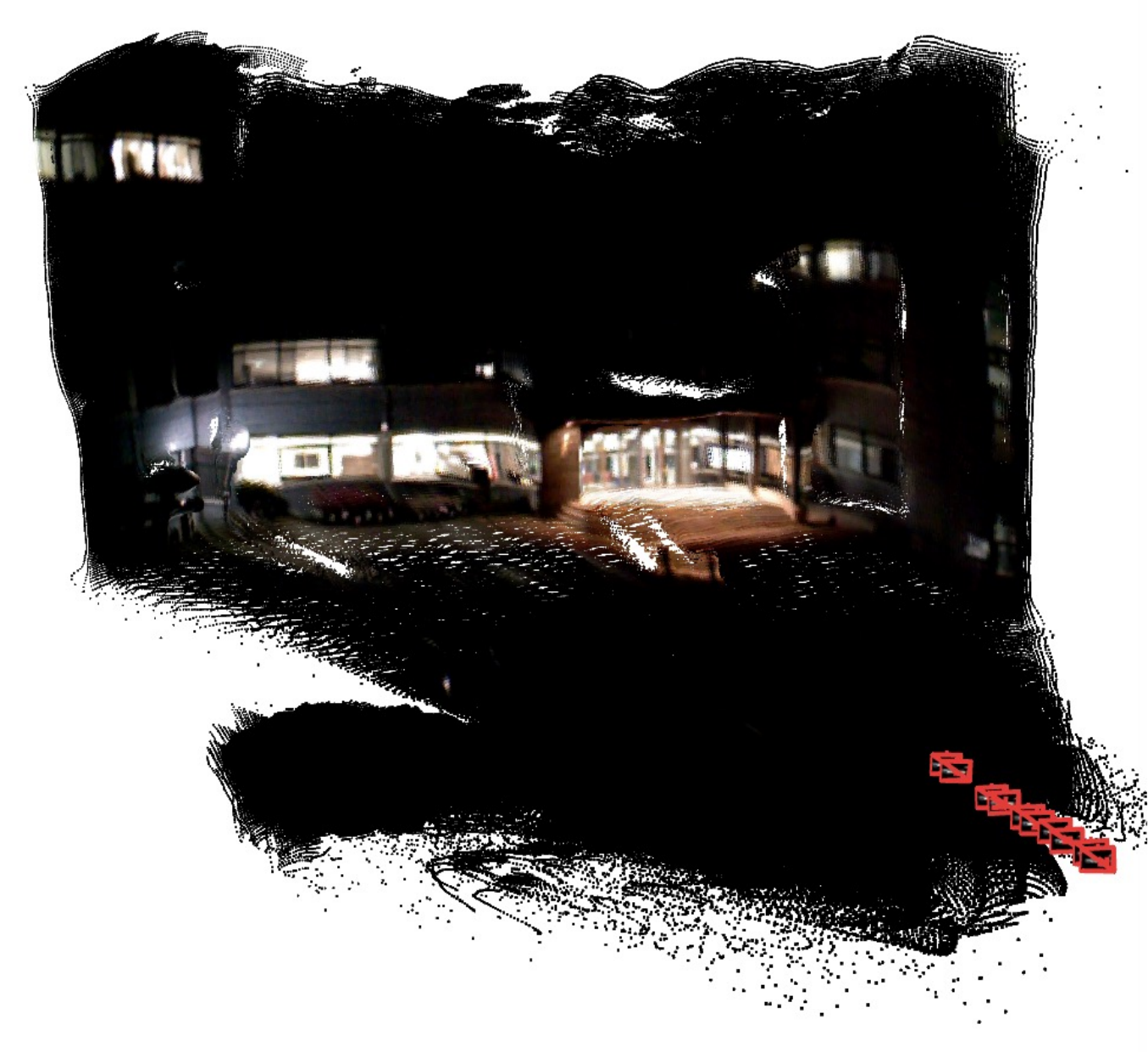}%
                    }%
                }%
            \end{minipage}%
            \hfill
            \begin{minipage}[c]{0.34\linewidth}
                \centering
                \scalebox{1}[1.10]{%
                    \includegraphics[
                        width=\linewidth
                    ]{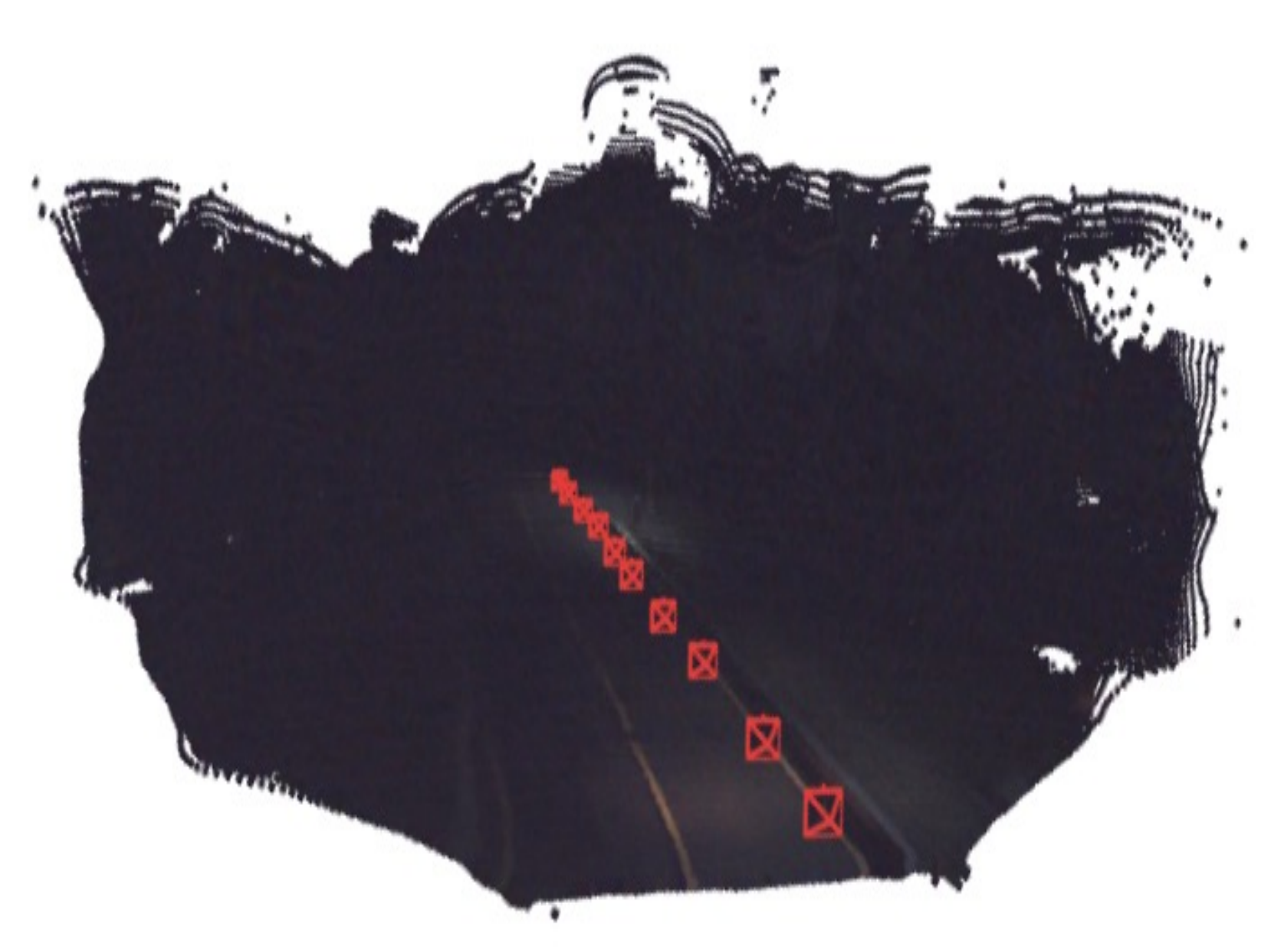}%
                }%
            \end{minipage}%
            \hfill
            \begin{minipage}[c]{0.32\linewidth}
                \centering
                \makebox[\linewidth][c]{%
                    \includegraphics[
                        width=1\linewidth,
                        height=0.755\linewidth
                    ]{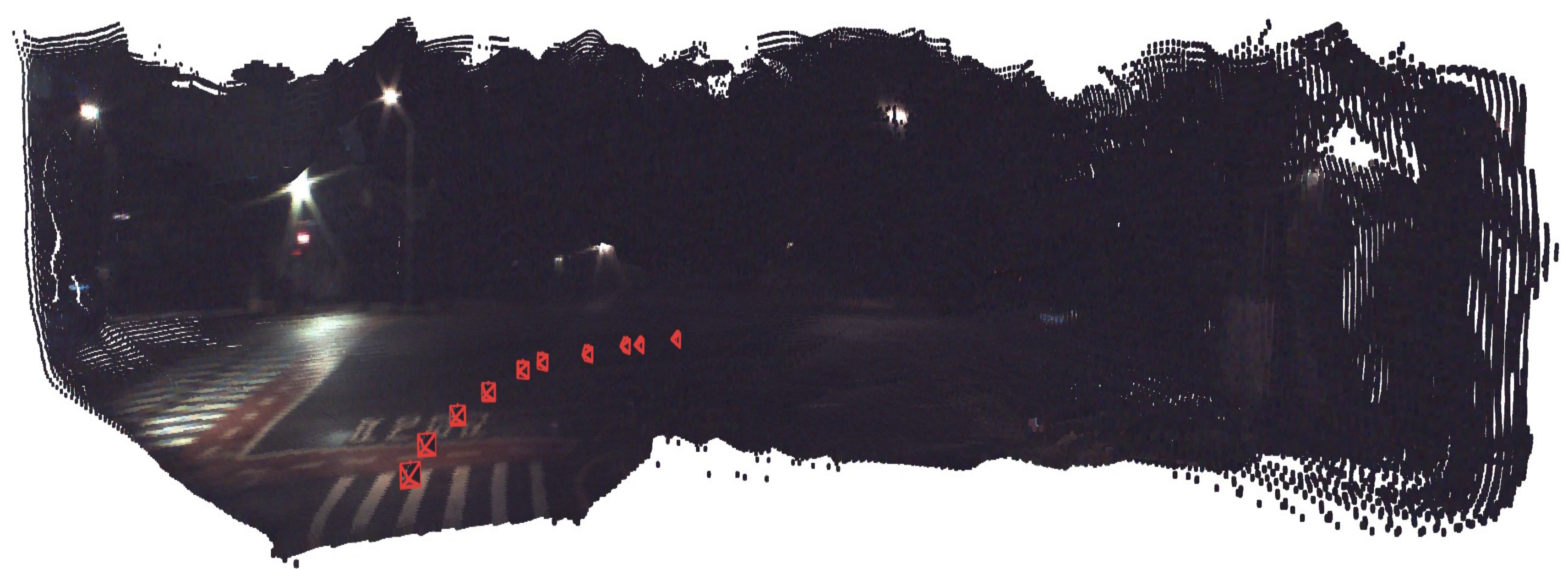}%
                }%
            \end{minipage}%
        \end{minipage}%
        \\ \noalign{\vskip 2.em}

        & \text{SEAR} & \textbf{DarkVGGT (Ours)} \\[0.2em]

        \makebox[0.03\textwidth][l]{%
            \rotatebox[origin=c]{90}{\text{RGB-T}}%
        }%
        &
        \begin{minipage}[c]{0.475\textwidth}
            \noindent
            \begin{minipage}[c]{0.32\linewidth}
                \centering
                \makebox[\linewidth][c]{%
                    \scalebox{1}[1]{%
                        \includegraphics[
                            width=\linewidth
                        ]{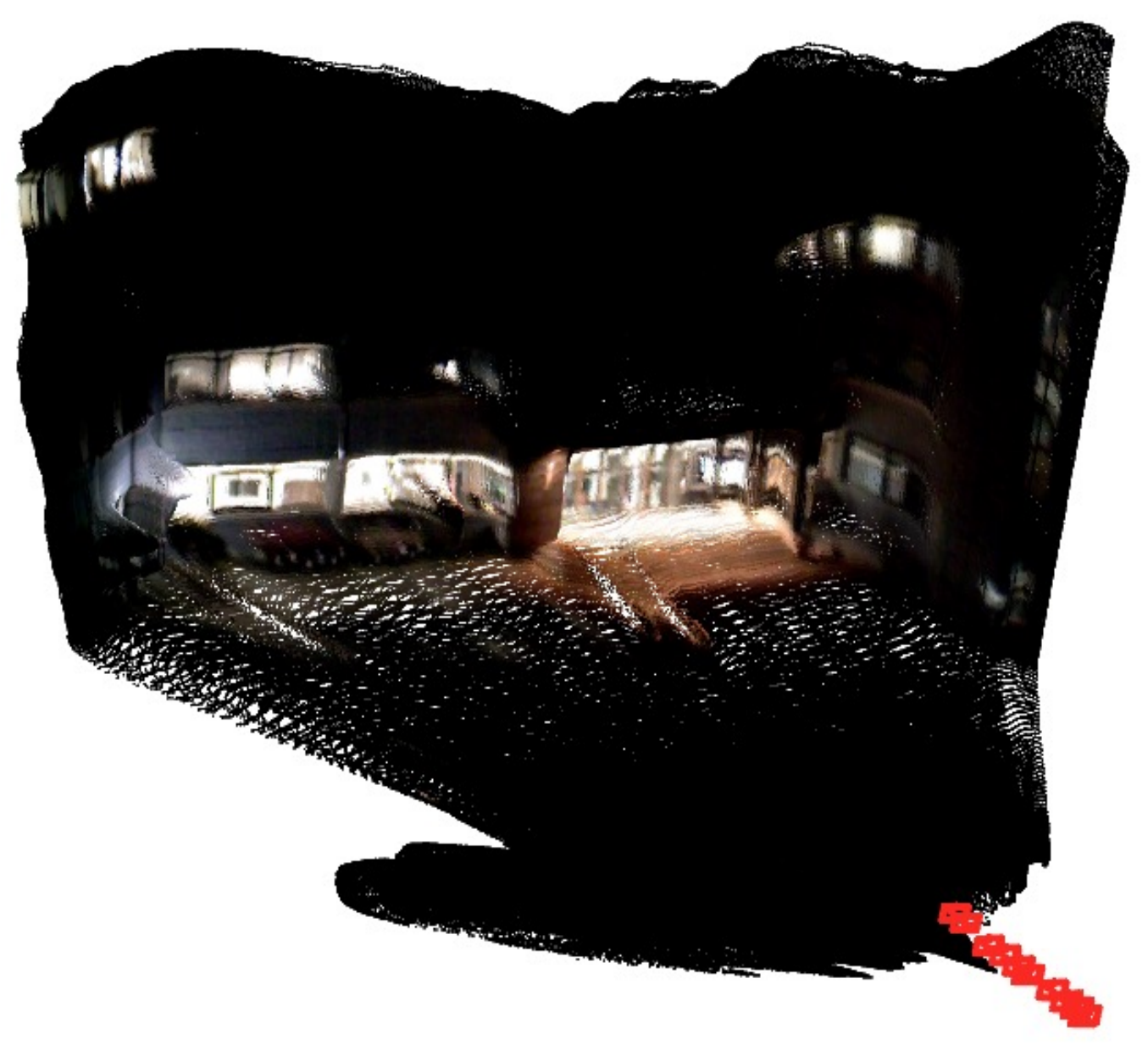}%
                    }%
                }%
            \end{minipage}%
            \hfill
            \begin{minipage}[c]{0.34\linewidth}
                \centering
                \scalebox{1}[1]{%
                    \includegraphics[
                        width=\linewidth
                    ]{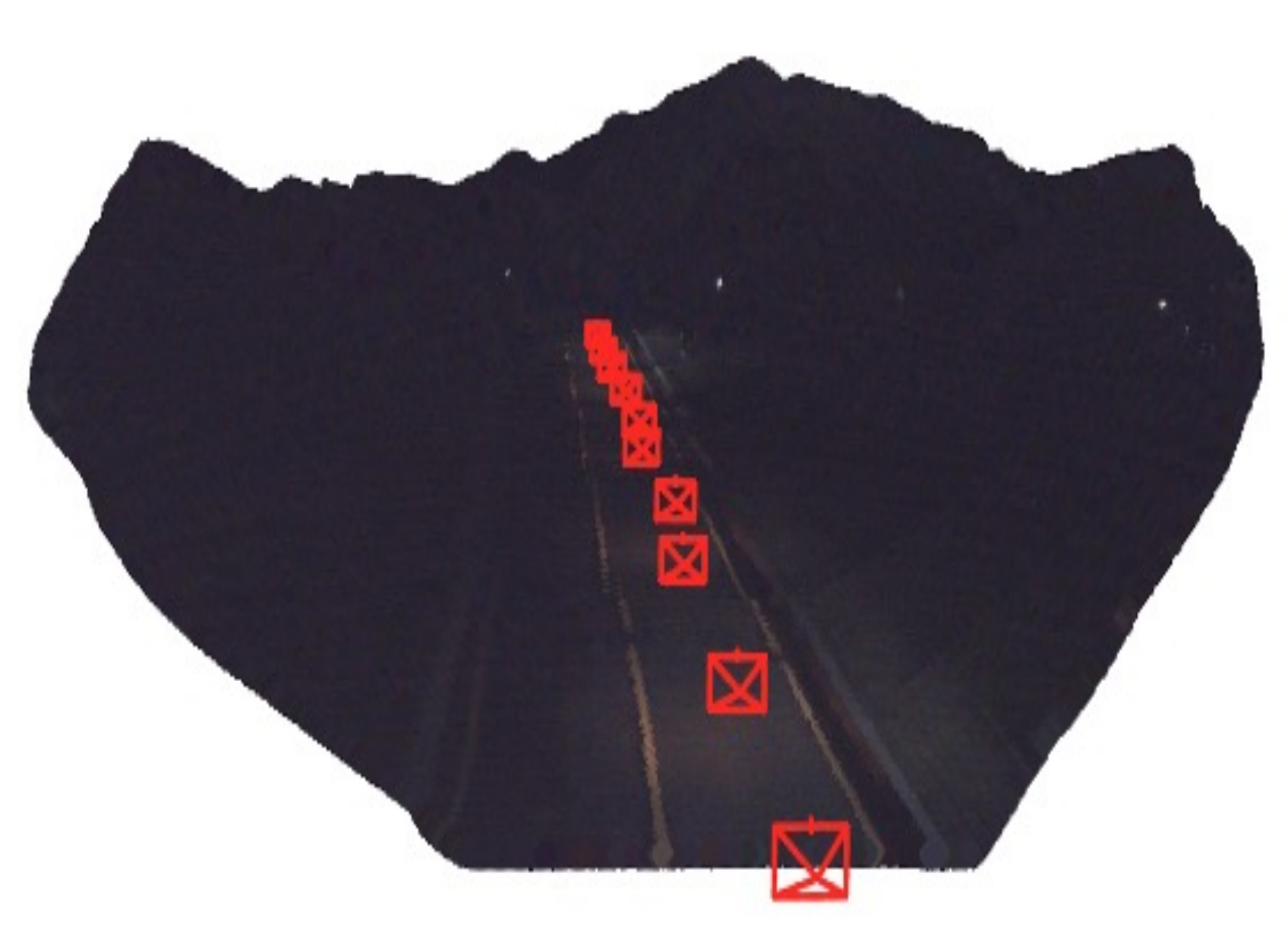}%
                }%
            \end{minipage}%
            \hfill
            \begin{minipage}[c]{0.32\linewidth}
                \centering
                \makebox[\linewidth][c]{%
                    \includegraphics[
                        width=1\linewidth,
                        height=0.755\linewidth
                    ]{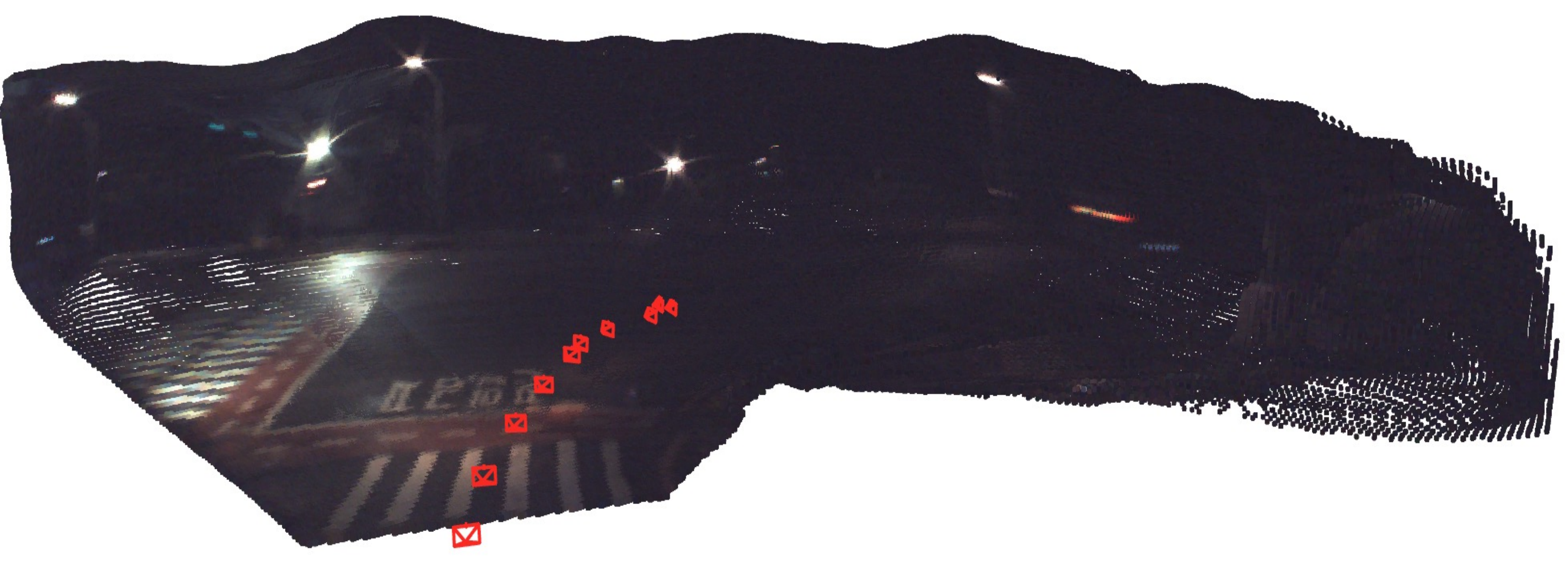}%
                }%
            \end{minipage}%
            \par\vspace{0.1em}
            \noindent
            \begin{minipage}[c]{0.32\linewidth}
                \centering
                \makebox[\linewidth][c]{%
                    \scalebox{1}[1]{%
                        \includegraphics[
                            width=\linewidth
                        ]{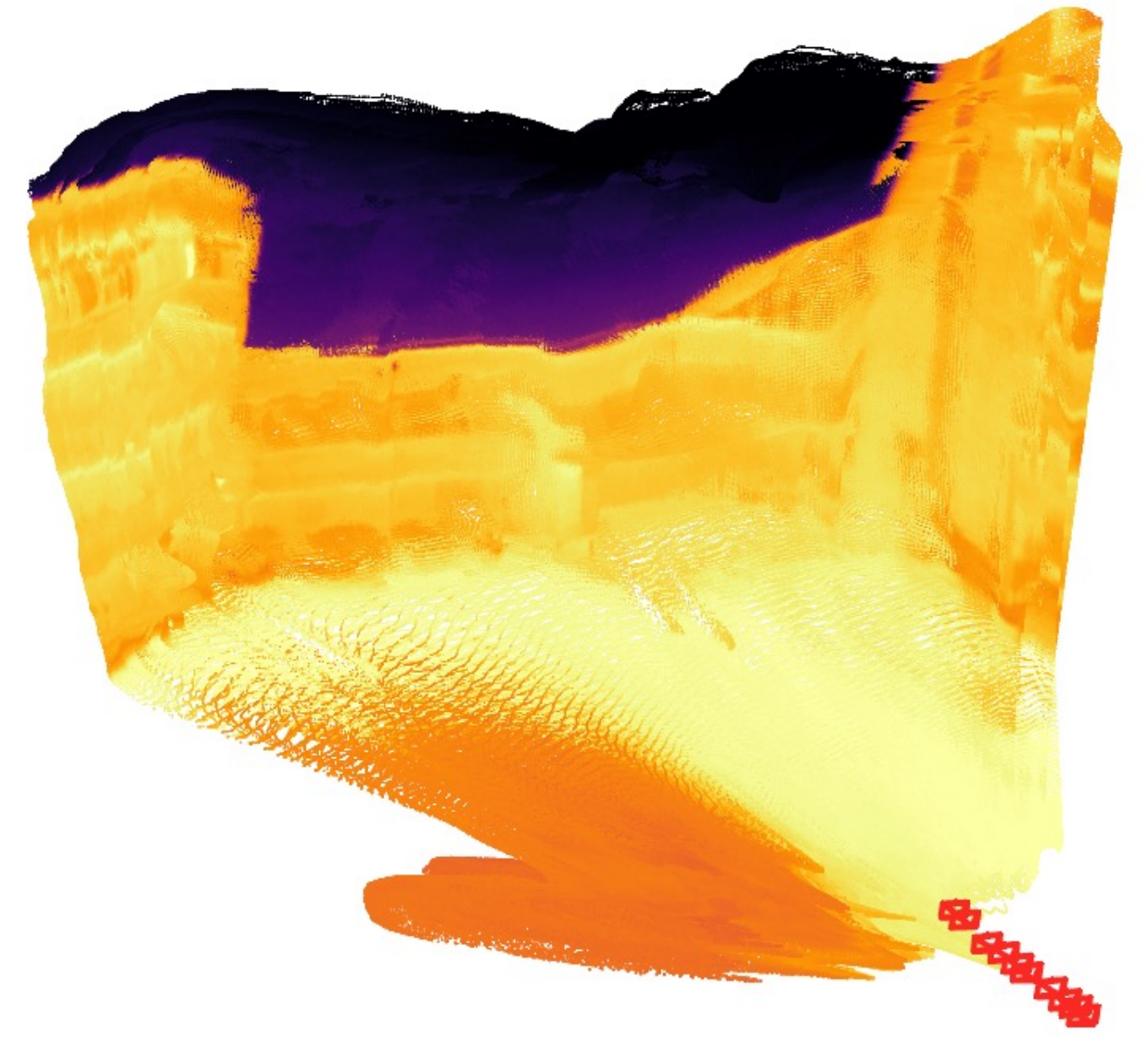}%
                    }%
                }%
            \end{minipage}%
            \hfill
            \begin{minipage}[c]{0.34\linewidth}
                \centering
                \scalebox{1}[1]{%
                    \includegraphics[
                        width=\linewidth
                    ]{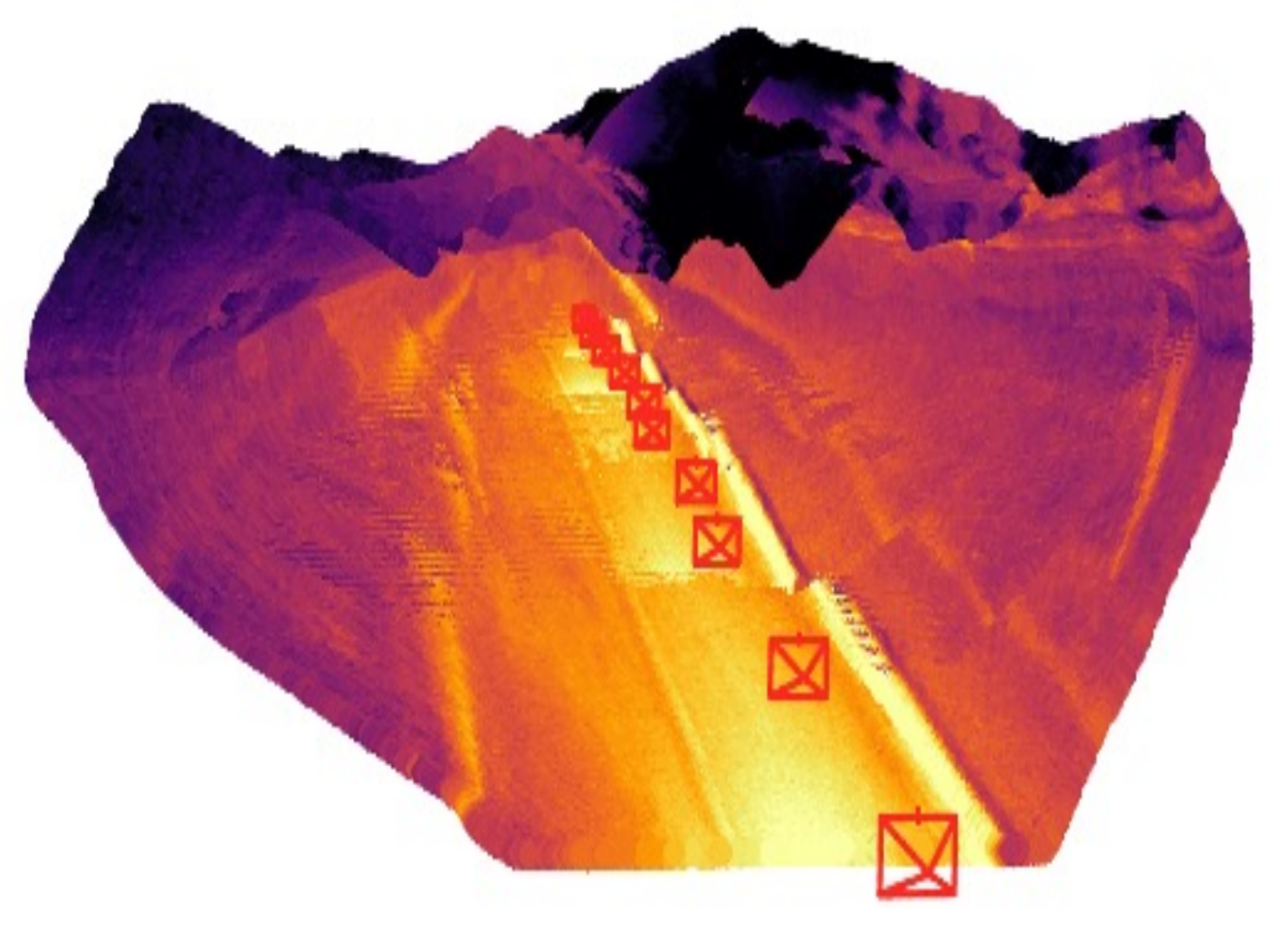}%
                }%
            \end{minipage}%
            \hfill
            \begin{minipage}[c]{0.32\linewidth}
                \centering
                \makebox[\linewidth][c]{%
                    \includegraphics[
                        width=1\linewidth,
                        height=0.755\linewidth
                    ]{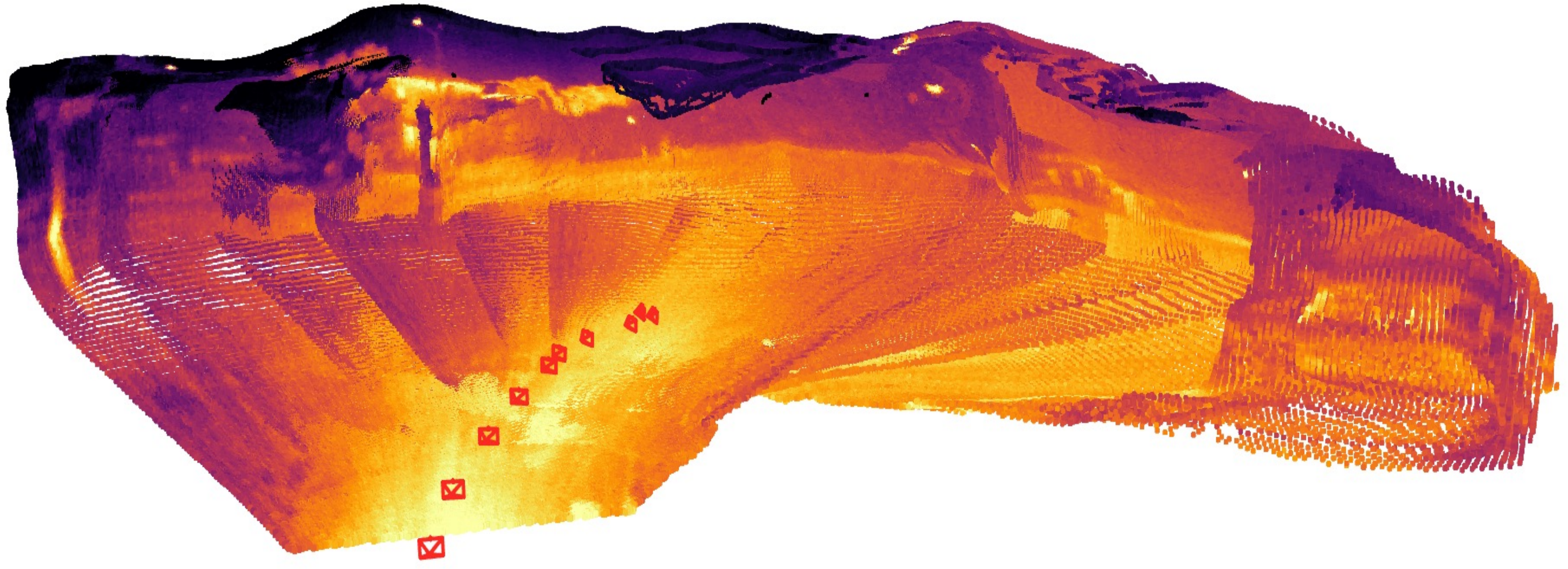}%
                }%
            \end{minipage}%
        \end{minipage}%
        &
        \begin{minipage}[c]{0.475\textwidth}
            \noindent
            \begin{minipage}[c]{0.32\linewidth}
                \centering
                \makebox[\linewidth][c]{%
                    \scalebox{1}[1]{%
                        \includegraphics[
                            width=\linewidth
                        ]{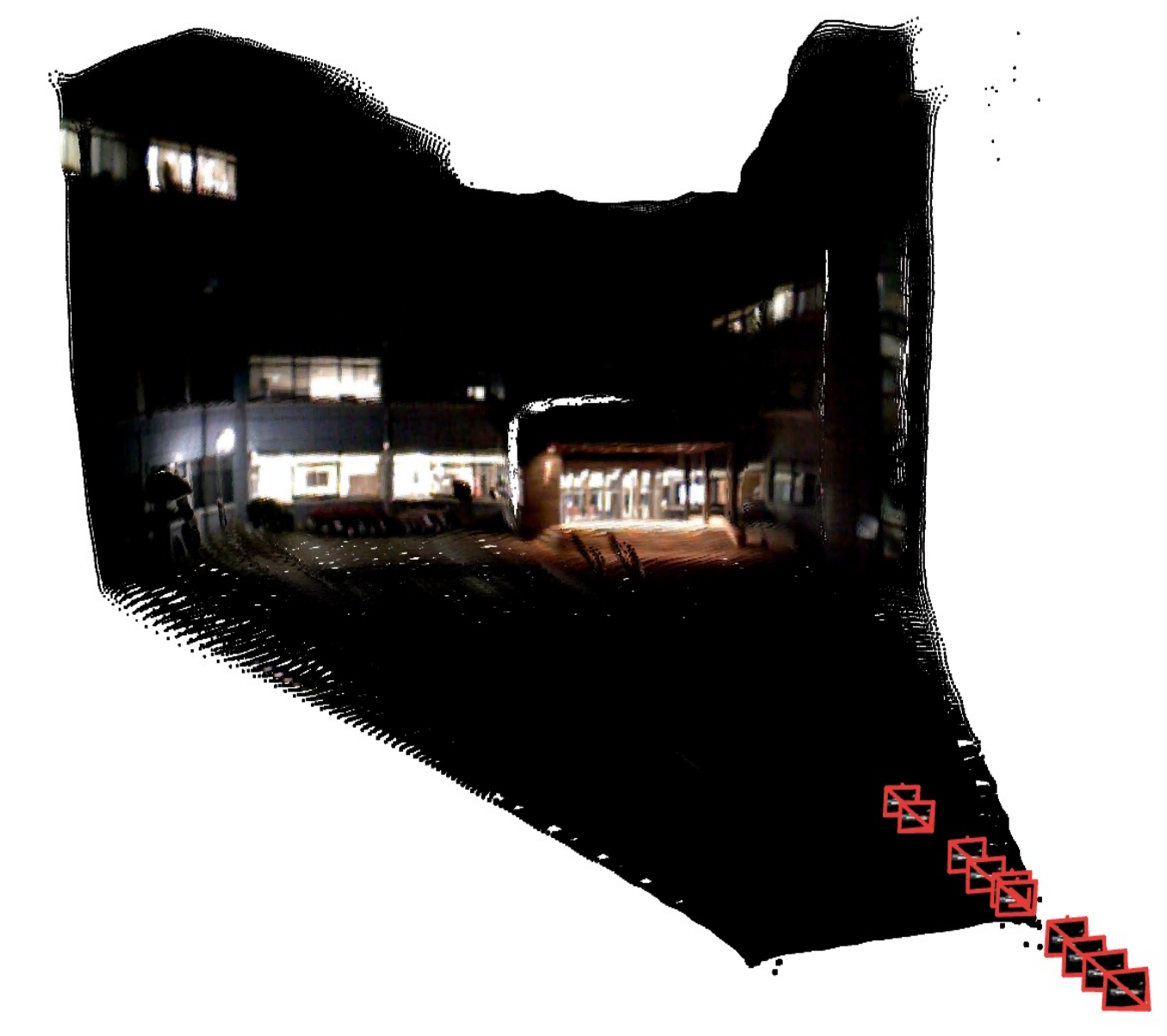}%
                    }%
                }%
            \end{minipage}%
            \hfill
            \begin{minipage}[c]{0.34\linewidth}
                \centering
                \scalebox{1}[1]{%
                    \includegraphics[
                        width=\linewidth
                    ]{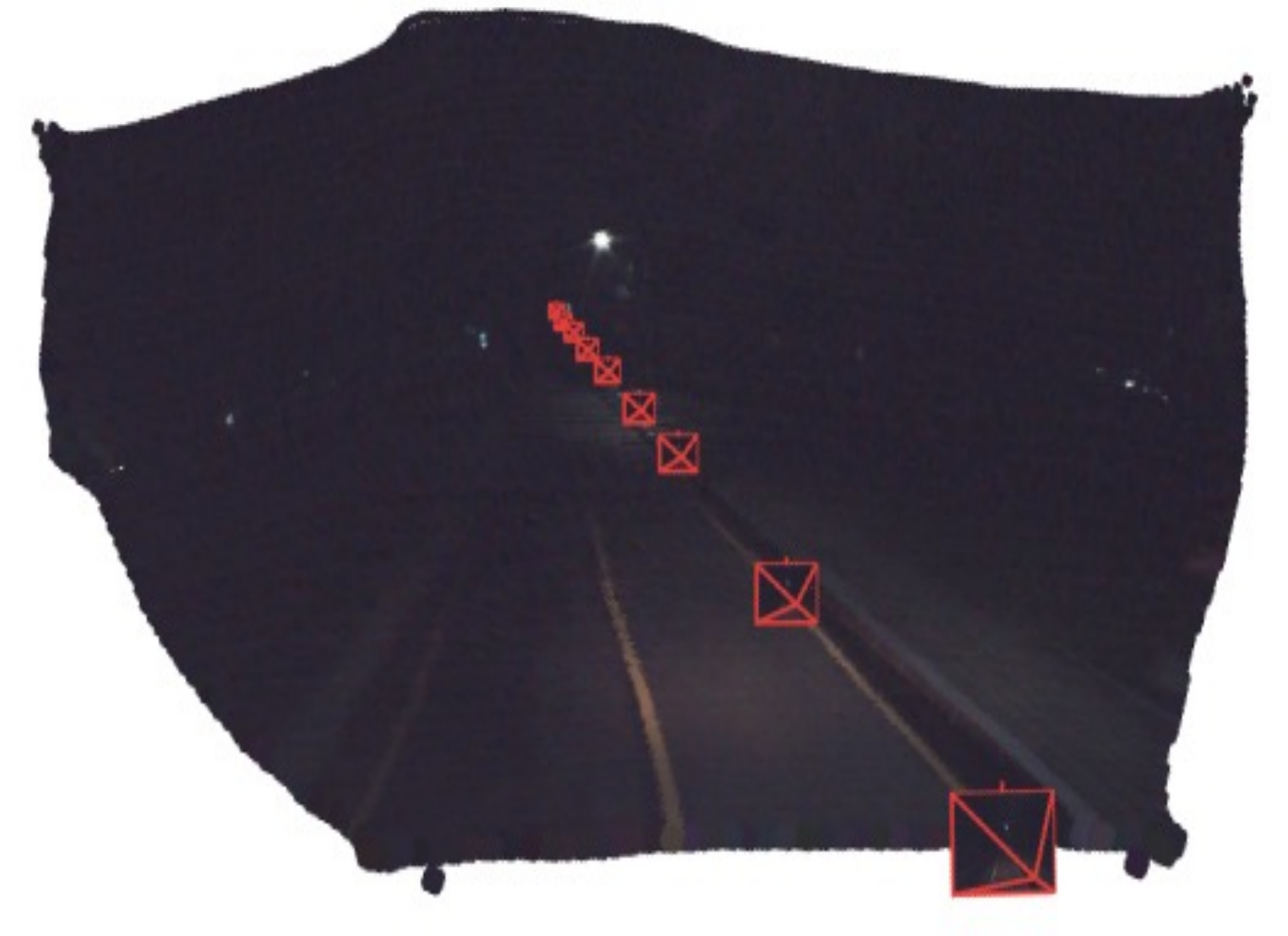}%
                }%
            \end{minipage}%
            \hfill
            \begin{minipage}[c]{0.32\linewidth}
                \centering
                \makebox[\linewidth][c]{%
                    \includegraphics[
                        width=1\linewidth,
                        height=0.755\linewidth
                    ]{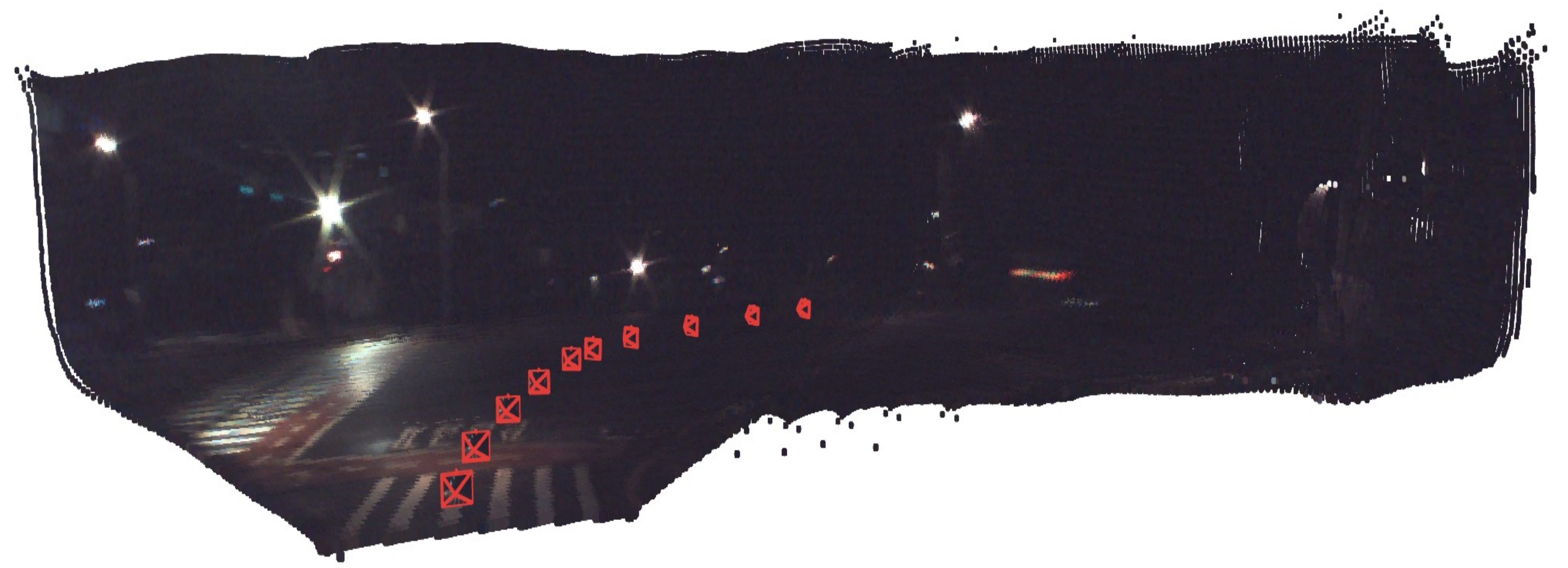}%
                }%
            \end{minipage}%
            \par\vspace{0.1em}
            \noindent
            \begin{minipage}[c]{0.32\linewidth}
                \centering
                \makebox[\linewidth][c]{%
                    \scalebox{1}[1]{%
                        \includegraphics[
                            width=\linewidth
                        ]{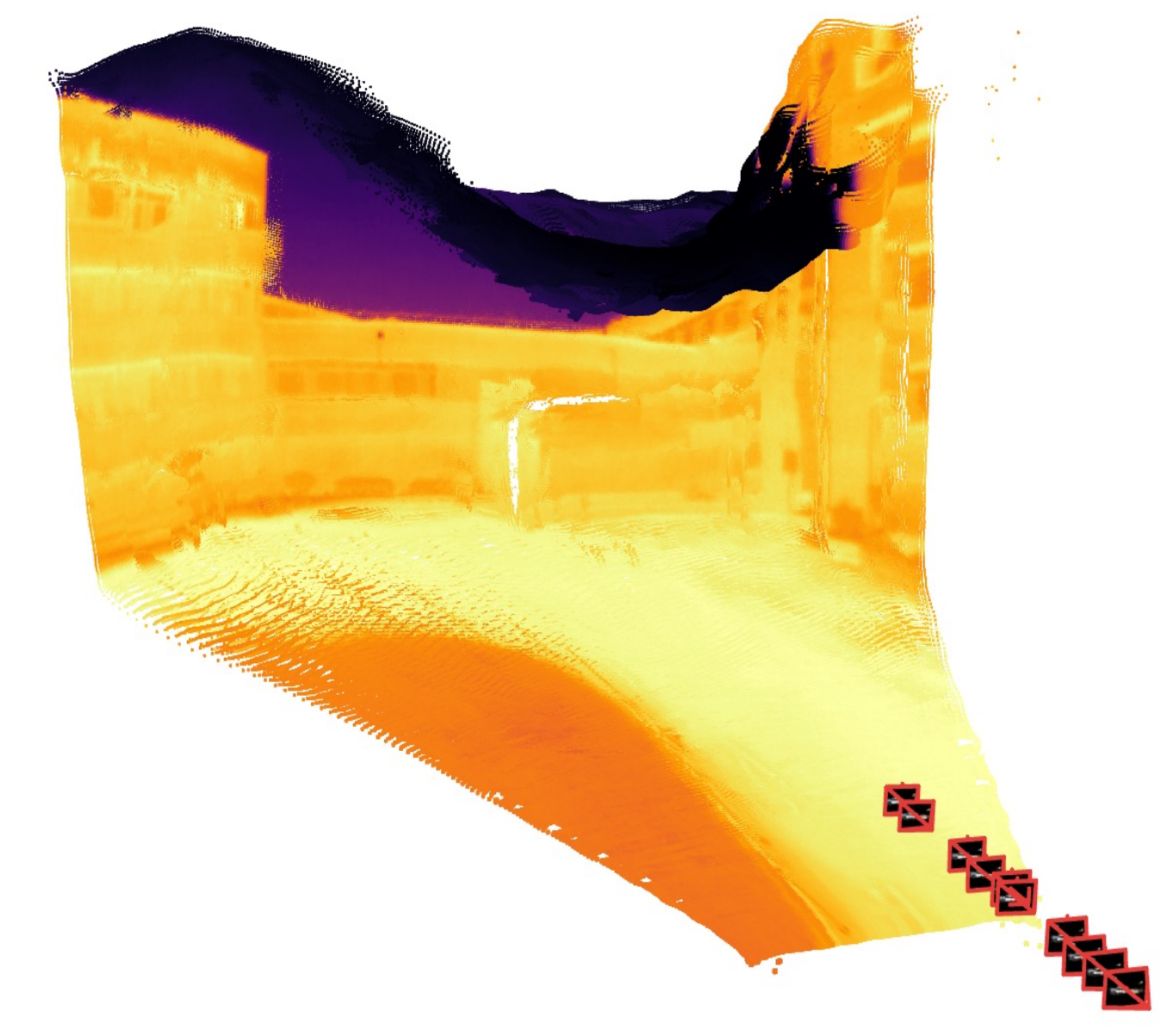}%
                    }%
                }%
            \end{minipage}%
            \hfill
            \begin{minipage}[c]{0.34\linewidth}
                \centering
                \scalebox{1}[1]{%
                    \includegraphics[
                        width=\linewidth
                    ]{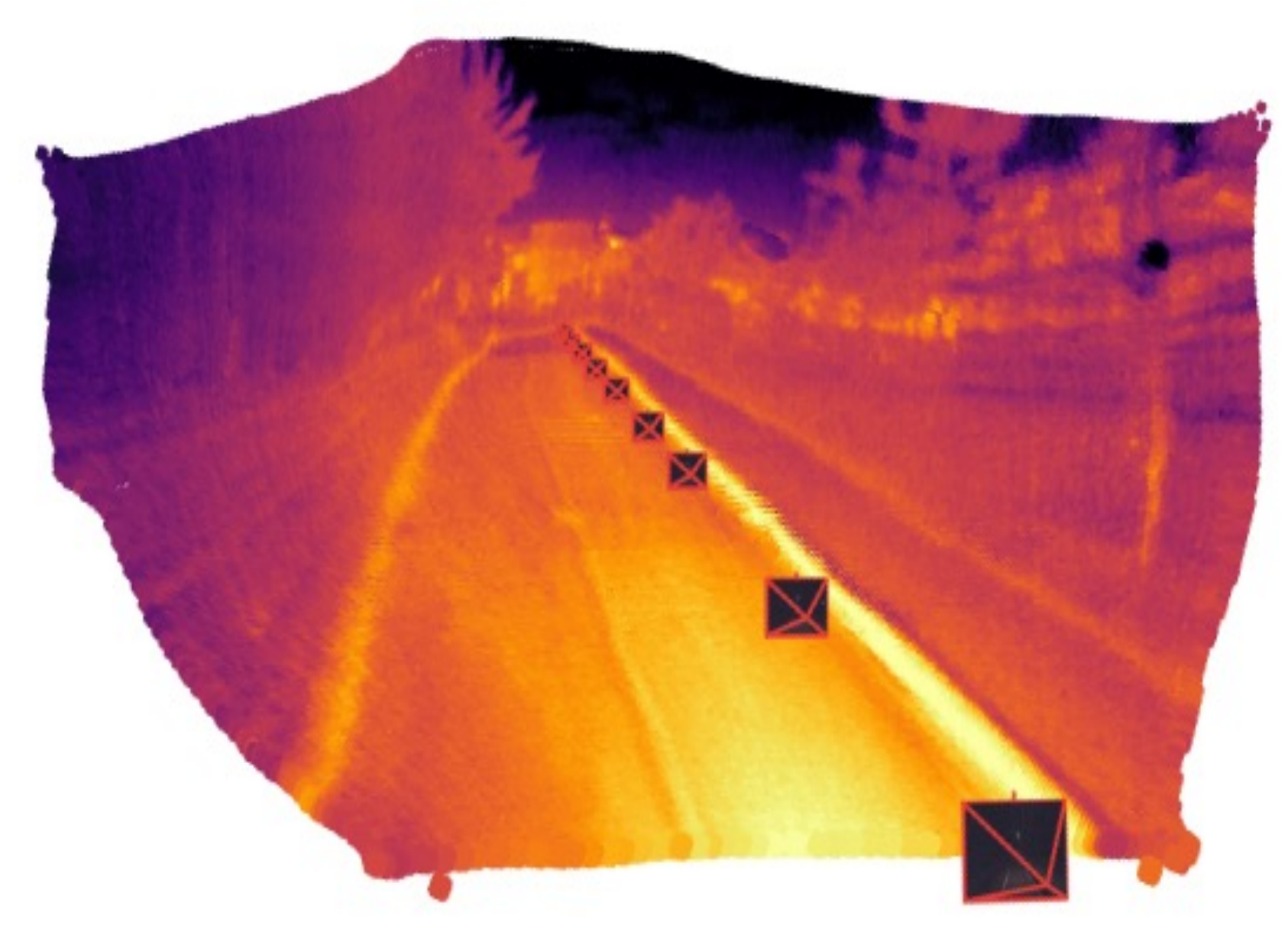}%
                }%
            \end{minipage}%
            \hfill
            \begin{minipage}[c]{0.32\linewidth}
                \centering
                \makebox[\linewidth][c]{%
                    \includegraphics[
                        width=1\linewidth,
                        height=0.755\linewidth
                    ]{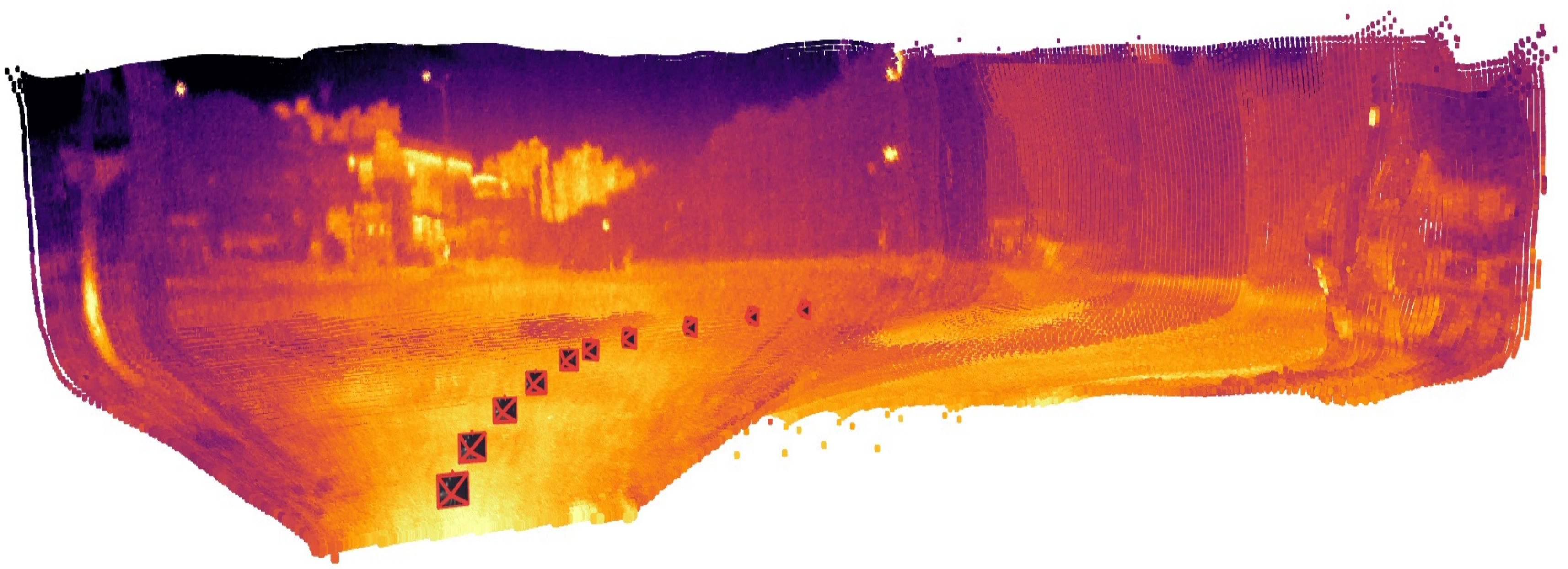}%
                }%
            \end{minipage}%
        \end{minipage}%
    \end{tabular}

    \caption{Qualitative comparison of nighttime 3D geometry estimation across Dark3R, VGGT, SEAR, and DarkVGGT. Red cameras indicate predicted poses. For RGB-T methods, we show RGB-textured and thermal-intensity-colored point-cloud reconstructions.}
    \label{fig:darkvggt_fig3}
\end{figure*}

\subsection{Training}
\paragraph{Thermal dropout and RGB-prior preservation.}
To reduce over-reliance on thermal cues and preserve the pretrained RGB prior, we apply thermal-branch dropout during training on batches whose clips all come from RGB-dominant scenes (see Appendix~\ref{app:hyperparam}). For such batches, the thermal inputs $\{I_s^{\thr}\}_{s=1}^{S}$ are dropped with a fixed probability, so the model performs an RGB-only forward pass. On these dropout steps, we additionally regularize the adapted model toward a frozen base-VGGT reference evaluated on the same RGB frames: $\mathcal{L}_{\mathrm{drop}}=
\|\hat{D}_{\mathrm{rgb}}-\hat{D}^{0}_{\mathrm{rgb}}\|_2^2 + \|\hat{P}_{\mathrm{rgb}}-\hat{P}^{0}_{\mathrm{rgb}}\|_2^2$,
where $(\hat{D}_{\mathrm{rgb}},\hat{P}_{\mathrm{rgb}})$ and $(\hat{D}^{0}_{\mathrm{rgb}},\hat{P}^{0}_{\mathrm{rgb}})$ denote RGB-only depth and pose predictions from the adapted student and the frozen base-VGGT reference, respectively. This term directly counteracts the daylight tax by anchoring the RGB-only forward of the adapted model to its pretrained behavior.

\paragraph{Training losses.}
DarkVGGT optimizes
\begin{equation}
\mathcal{L}
=
\mathcal{L}_{\mathrm{VGGT}}
+\mathcal{L}_{\mathrm{ctff}}
+\mathcal{L}_{\mathrm{gstr}}
+\mathcal{I}_{\mathrm{drop}}\,\lambda_{\mathrm{drop}}\,\mathcal{L}_{\mathrm{drop}},
\label{eq:loss}
\end{equation}
where $\mathcal{I}_{\mathrm{drop}}$ equals 1 on thermal-dropout steps and 0 otherwise, and $\mathcal{L}_{\mathrm{VGGT}}$ follows VGGT~\citep{wang2025vggt}. The terms $\mathcal{L}_{\mathrm{ctff}}$ and $\mathcal{L}_{\mathrm{gstr}}$ are weighted sums of the auxiliary losses from the complementary thermal feature factorization module described in Section~\ref{sec:method_phys} and the geometry-shared thermal routing module described in Section~\ref{sec:method_gstr}, respectively.

\begin{table*}[t]
\centering
\small
\caption{Depth estimation across low-visibility scenes. DarkVGGT is compared with feed-forward 3D reconstruction baselines, reporting quantitative depth metrics to evaluate robustness under severe illumination degradation. $\dagger$ denotes a model fine-tuned under our training setup. \protect\colorbox{red!25}{\strut 1st}\,
\protect\colorbox{orange!25}{\strut 2nd}\,
\protect\colorbox{yellow!35}{\strut 3rd}.}
\label{tab:depth_main}
\setlength{\tabcolsep}{5pt}
\resizebox{\textwidth}{!}{%
\begin{tabular}{l *{3}{c} *{3}{c} *{3}{c}}
\toprule
 & \multicolumn{3}{c}{ViViD++}
 & \multicolumn{3}{c}{STheReO}
 & \multicolumn{3}{c}{Dark3R} \\
\cmidrule(lr){2-4}\cmidrule(lr){5-7}\cmidrule(lr){8-10}
Method
 & AbsRel$\downarrow$ & $\delta\!<\!1.25\uparrow$ & RMSE$_{\log}\!\downarrow$
 & AbsRel$\downarrow$ & $\delta\!<\!1.25\uparrow$ & RMSE$_{\log}\!\downarrow$
 & AbsRel$\downarrow$ & $\delta\!<\!1.25\uparrow$ & RMSE$_{\log}\!\downarrow$ \\
\midrule
\multicolumn{10}{l}{\textbf{(A) RGB-only}} \\
DUSt3R~\cite{wang2024dust3r}
 & 0.3765 & 0.4883 & 0.4008
 & 0.4759 & 0.3146 & 0.6437
 & 0.2118 & 0.6391 & 0.2562 \\
MASt3R~\cite{leroy2024grounding}
 & 0.2685 & 0.6121 & 0.3132
 & 0.5093 & 0.3418 & 0.6619
 & 0.2004 & 0.6933 & 0.2496 \\
Dark3R~\cite{guo2026dark3r}
 & 0.3036 & 0.5739 & 0.3392
 & 0.4843 & 0.3404 & 0.6437
 & \thirdcell{0.1510} & \thirdcell{0.7914} & \thirdcell{0.1960} \\
DepthAnything3~\cite{depthanything3}
 & 0.2552 & 0.6239 & 0.2990
 & 0.3145 & 0.4746 & 0.5246
 & 0.2054 & 0.6528 & 0.2489 \\
VGGT~\cite{wang2025vggt}
 & 0.2746 & 0.5874 & 0.3234
 & 0.2950 & 0.5156 & 0.4567
 & 0.2433 & 0.6095 & 0.2861 \\
\midrule
\multicolumn{10}{l}{\textbf{(B) RGB-T}} \\
SEAR~\cite{skorokhodov2026sear}
 & \thirdcell{0.1956} & \thirdcell{0.7502} & \thirdcell{0.2421}
 & \thirdcell{0.2370} & \thirdcell{0.6143} & \thirdcell{0.4147}
 & 0.1703 & 0.7058 & 0.2151 \\
SEAR$^{\dagger}$~\cite{skorokhodov2026sear}
 & \secondcell{0.1806} & \secondcell{0.7670} & \secondcell{0.2378}
 & \secondcell{0.1780} & \secondcell{0.7323} & \secondcell{0.3399}
 & \secondcell{0.1305} & \secondcell{0.8274} & \secondcell{0.1696} \\
DarkVGGT (ours)
 & \bestcell{0.1274} & \bestcell{0.8692} & \bestcell{0.1970}
 & \bestcell{0.1600} & \bestcell{0.7651} & \bestcell{0.3077}
 & \bestcell{0.1246} & \bestcell{0.8387} & \bestcell{0.1651} \\
\bottomrule
\end{tabular}%
}
\end{table*}

\section{Experiments}
\label{sec:experiments}
\paragraph{Implementation details.}
We initialize from the publicly released VGGT-1B weights~\citep{wang2025vggt} and keep the backbone frozen. Trainable parameters comprise LoRA adapters ($r=64$, $\alpha=128$) on the AA blocks, a thermal camera token and RoPE offset, the CTFF allocation weights and fusion gates, and the shared--private decomposition with ranks $(d_s,d_p,d_{\mathrm{mi}})=(96,192,128)$ active in the last $k=8$ AA blocks. We train DarkVGGT on four NVIDIA A100 GPUs at a resolution of $518\times518$ using bfloat16 AMP for $10$ epochs with $200$ optimization steps per epoch and thermal dropout probability $p=0.5$. We optimize the model with AdamW using a learning rate of $5\times10^{-5}$, a weight decay of $10^{-2}$, and cosine decay, with the remaining hyperparameters provided in Appendix~\ref{app:hyperparam}.

Additional controlled input and fusion comparisons and the scope of the learned supervision prior are provided in Appendices~\ref{app:controlled_baselines} and~\ref{app:supervision_prior}.

The RGB-only baselines are evaluated zero-shot. We report both off-the-shelf SEAR and SEAR$^{\dagger}$, fine-tuned with the same paired splits, depth supervision, input resolution, VGGT-1B initialization, optimization schedule, and training budget as DarkVGGT. Comparisons use the stronger SEAR variant for each dataset and metric, since fine-tuning does not uniformly improve pose estimation.

\paragraph{Datasets.}
We evaluate DarkVGGT on two real multi-sensor RGB-T datasets, ViViD++\citep{lee2022vivid++} and STheReO\citep{yun2022sthereo}, and report Dark3R~\citep{guo2026dark3r} as a complementary pseudo-thermal low-light evaluation.
Since Dark3R does not provide measured thermal images, we synthesize pseudo-thermal inputs with ThermalGen~\citep{xiao2025thermalgen} and treat it separately from real RGB-T benchmarks.
For ViViD++ and STheReO, we supervise dense depth prediction using LiDAR-derived dense depth maps refined with Lingbot-Depth~\citep{tan2026masked}, while for Dark3R, we use the released canonical dense depth maps. For evaluation, we use \texttt{indoor\_aggresive\_dark}, \texttt{indoor\_unstable\_dark}, and \texttt{outdoor\_robust\_night2} from ViViD++, \texttt{kaist\_evening} and \texttt{valley\_evening} from STheReO, and \texttt{dark\_chapel} and \texttt{dark\_kitchen} from Dark3R as test sequences.
To assess RGB-prior preservation under standard light conditions, we evaluate RGB-only performance on ETH3D~\citep{schops2017multi} and ScanNet++~\citep{yeshwanth2023scannet++}.
Each test sequence is split into non-overlapping 50 frame chunks with 10 frames uniformly sampled per chunk following VGGT~\citep{wang2025vggt}, except for Dark3R dataset, where we use a fixed stride of $20$ across the full scene.
Detailed data preprocessing and train-test sequence splits are provided in Appendix~\ref{app:rgbt_processing}. Dark3R frame-sampling variance is analyzed in Appendix~\ref{app:dark3r_variability}.

\paragraph{Metrics.}
We evaluate depth and pose estimation performance following the protocol used in prior work~\citep{wang2025vggt,zhang2025flare}.
For depth estimation, we report AbsRel, $\delta<1.25$, and $\mathrm{RMSE}_{\log}$, which measure relative depth error, threshold-based accuracy, and logarithmic depth deviation, respectively. We use raw LiDAR measurements as ground-truth depth for ViViD++~\citep{lee2022vivid++} and STheReO~\citep{yun2022sthereo}, and the released canonical depth maps for Dark3R~\citep{guo2026dark3r}. For pose estimation, we report $\mathrm{RRA}$, $\mathrm{RTA}$, and $\mathrm{AUC}$, corresponding to relative rotation accuracy, relative translation accuracy, and the area under the cumulative pose-accuracy curve, respectively.

\paragraph{Depth estimation performance comparison.}
Table~\ref{tab:depth_main} compares the depth estimation performance of DarkVGGT with RGB-only and RGB-T feed-forward 3D reconstruction models across three low-visibility datasets.
DarkVGGT achieves the lowest numerical AbsRel and $\mathrm{RMSE}_{\log}$, as well as the highest $\delta<1.25$ across all three datasets.
Compared with fine-tuned SEAR~\citep{skorokhodov2026sear}, which achieves better depth scores than its off-the-shelf variant on all three benchmarks, DarkVGGT reduces average AbsRel from 0.1630 to 0.1373, increases $\delta<1.25$ from 0.7756 to 0.8243, and lowers $\mathrm{RMSE}_{\log}$ from 0.2491 to 0.2232. DarkVGGT and fine-tuned SEAR achieve comparable depth performance on Dark3R dataset.
Against the RGB-only model Dark3R~\citep{guo2026dark3r}, DarkVGGT yields lower AbsRel on all three benchmarks, with the largest gains on ViViD++ and STheReO, while also reducing $\mathrm{RMSE}_{\log}$ across the board.
On the ViViD++ and STheReO datasets, DepthAnything3~\citep{depthanything3} and VGGT~\citep{wang2025vggt} obtain the best numerical scores among RGB-only models, respectively, yet both remain substantially behind RGB-T methods. This suggests that thermal cues provide an effective inductive bias when adapting feed-forward 3D reconstruction models to dark scenes. In particular, compared with our baseline model, VGGT, DarkVGGT reduces the average AbsRel and $\mathrm{RMSE}_{\log}$ over the three datasets by 0.1337 and 0.1322, respectively, while improving the average $\delta<1.25$ by 0.2535.

These results support complementary thermal feature factorization and gated shared-subspace correction for multimodal adaptation and geometric prediction in low-visibility scenes.
As shown in Figure~\ref{fig:darkvggt_fig3}, DarkVGGT improves depth estimation in nighttime scenes and reconstructs more accurate dense 3D point clouds than existing RGB-only models, including Dark3R and VGGT, as well as the RGB-T model SEAR.

\begin{table*}[t]
\centering
\small
\caption{Camera pose estimation on low-visibility RGB-T scenes. We compare DarkVGGT with feed-forward 3D reconstruction baselines and report pose metrics under weak geometric correspondences.}
\label{tab:pose_main}
\setlength{\tabcolsep}{3pt}
\resizebox{\textwidth}{!}{%
\begin{tabular}{l *{3}{c} *{3}{c} *{3}{c}}
\toprule
 & \multicolumn{3}{c}{ViViD++}
 & \multicolumn{3}{c}{STheReO}
 & \multicolumn{3}{c}{Dark3R} \\
\cmidrule(lr){2-4}\cmidrule(lr){5-7}\cmidrule(lr){8-10}
Method
 & $\text{RRA}@30\uparrow$ & $\text{RTA}@30\uparrow$ & $\text{AUC}@30\uparrow$
 & $\text{RRA}@30\uparrow$ & $\text{RTA}@30\uparrow$ & $\text{AUC}@30\uparrow$
 & $\text{RRA}@30\uparrow$ & $\text{RTA}@30\uparrow$ & $\text{AUC}@30\uparrow$ \\
\midrule
\multicolumn{10}{l}{\textbf{(A) RGB-only}} \\
DUSt3R~\cite{wang2024dust3r}
 & 0.72 & 0.22 & 6.4
 & 0.88 & 0.43 & 27.0
 & 0.98 & 0.20 & 10.0 \\
MASt3R~\cite{leroy2024grounding}
 & 0.76 & 0.46 & 27.1
 & 0.98 & 0.80 & 71.3
 & \bestcell{1.00} & \thirdcell{0.89} & \secondcell{62.5} \\
Dark3R~\cite{guo2026dark3r}
 & 0.74 & 0.41 & 22.7
 & 0.99 & 0.79 & 69.8
 & \bestcell{1.00} & \secondcell{0.94} & \bestcell{68.0} \\
DepthAnything3~\cite{depthanything3}
 & 0.73 & 0.28 & 12.0
 & 0.99 & 0.85 & 76.6
 & 0.55 & 0.18 & 5.1 \\
VGGT~\cite{wang2025vggt}
 & 0.81 & \thirdcell{0.48} & 25.9
 & \bestcell{1.00} & \bestcell{0.99} & \bestcell{96.3}
 & 0.64 & 0.47 & 15.3 \\
\midrule
\multicolumn{10}{l}{\textbf{(B) RGB-T}} \\
SEAR~\cite{skorokhodov2026sear}
 & \secondcell{0.91} & \secondcell{0.59} & \secondcell{36.1}
 & \bestcell{1.00} & 0.97 & 88.5
 & 0.66 & 0.52 & 18.6 \\
SEAR$^{\dagger}$~\cite{skorokhodov2026sear}
 & \thirdcell{0.84} & 0.47 & \thirdcell{28.1}
 & \bestcell{1.00} & \secondcell{0.98} & \thirdcell{95.2}
 & 0.84 & 0.71 & 32.5 \\
DarkVGGT (ours)
 & \bestcell{0.98} & \bestcell{0.74} & \bestcell{46.0}
 & \bestcell{1.00} & \secondcell{0.98} & \secondcell{95.4}
 & \bestcell{1.00} & \bestcell{0.95} & \thirdcell{57.4} \\
\bottomrule
\end{tabular}%
}
\end{table*}

\begin{table*}[t]
\centering
\small
\caption{Depth and pose estimation on well-lit ETH3D and ScanNet++ scenes. We evaluate RGB-only performance to examine whether DarkVGGT preserves the pretrained VGGT geometry prior when thermal guidance is unavailable.}
\label{tab:well_lit}
\setlength{\tabcolsep}{6pt}
\resizebox{\textwidth}{!}{%
\begin{tabular}{l *{4}{c} *{4}{c}}
\toprule
 & \multicolumn{4}{c}{\textbf{ETH3D}}
 & \multicolumn{4}{c}{\textbf{ScanNet++}} \\
\cmidrule(lr){2-5}\cmidrule(lr){6-9}
Method
 & $\delta\!<\!1.25\uparrow$ & RMSE$_{\log}\!\downarrow$ & $\text{RRA}@30\uparrow$ & $\text{RTA}@30\uparrow$
 & $\delta\!<\!1.25\uparrow$ & RMSE$_{\log}\!\downarrow$ & $\text{RRA}@30\uparrow$ & $\text{RTA}@30\uparrow$ \\
\midrule
MASt3R~\cite{leroy2024grounding}
 & \thirdcell{0.9413} & \thirdcell{0.1496} & \thirdcell{0.67} & \thirdcell{0.68}
 & 0.8762 & 0.1890 & \thirdcell{0.89} & \thirdcell{0.90} \\

Dark3R~\cite{guo2026dark3r}
 & 0.8992 & 0.1739 & 0.63 & 0.65
 & \thirdcell{0.8889} & \thirdcell{0.1802} & 0.88 & 0.89 \\

VGGT~\cite{wang2025vggt}
 & \bestcell{0.9729} & \bestcell{0.0595} & \bestcell{1.00} & \bestcell{0.91}
 & \secondcell{0.9479} & \secondcell{0.1106} & \bestcell{0.98} & \bestcell{0.93} \\

\midrule
DarkVGGT (ours)
 & \secondcell{0.9584} & \secondcell{0.0803} & \bestcell{1.00} & \secondcell{0.88}
 & \bestcell{0.9608} & \bestcell{0.1016} & \bestcell{0.98} & \secondcell{0.92} \\
\bottomrule
\end{tabular}%
}
\end{table*}

\paragraph{Pose estimation performance comparison.}
Table~\ref{tab:pose_main} shows pose estimation performance comparing DarkVGGT with existing methods using $\mathrm{RRA}@30$, $\mathrm{RTA}@30$, and $\mathrm{AUC}@30$, where @30 denotes evaluation up to a $30^\circ$ angular threshold for relative rotation, translation-direction, and joint pose errors.
DarkVGGT achieves the strongest average pose performance across the three low-visibility datasets. Compared with the average of RGB-only baselines, DarkVGGT improves $\mathrm{RRA}@30$, $\mathrm{RTA}@30$, and $\mathrm{AUC}@30$ by 16.7\%, 59.1\%, and 66.8\%, respectively. Using the stronger SEAR~\citep{skorokhodov2026sear} variant for each dataset and metric, DarkVGGT raises the average $\mathrm{RRA}@30$, $\mathrm{RTA}@30$, and $\mathrm{AUC}@30$ from 0.917 to 0.993, from 0.760 to 0.890, and from 54.6 to 66.3, respectively. On STheReO, the two methods match in $\mathrm{RRA}@30$ and $\mathrm{RTA}@30$, with a small $\mathrm{AUC}@30$ difference of 95.2 versus 95.4. Relative to Dark3R~\citep{guo2026dark3r}, DarkVGGT also shows consistent average gains, with the largest improvement in $\mathrm{RTA}@30$. The overall pose scores on the STheReO dataset are higher than on the other two datasets, likely because its forward-driving scenes induce smaller relative pose errors.

Figure~\ref{fig:darkvggt_fig3} provides a qualitative view of these pose differences. Dark3R produces trajectories that collapse toward one side, leading to distorted dense point-cloud reconstructions. SEAR, as an RGB-T baseline, improves over VGGT~\citep{wang2025vggt} in 3D reconstruction quality but still exhibits locally warped camera trajectories. In contrast, DarkVGGT estimates more stable camera poses, supporting higher-quality 3D geometry prediction under nighttime conditions.

\paragraph{RGB-only performance in well-lit scenes.}
To assess whether DarkVGGT preserves the pretrained VGGT prior when thermal guidance is unavailable, we evaluate RGB-only performance on the well-lit ETH3D~\citep{schops2017multi} and ScanNet++~\citep{yeshwanth2023scannet++} datasets. As shown in Table~\ref{tab:well_lit}, DarkVGGT remains close to VGGT~\citep{wang2025vggt} overall and even improves some metrics. On ETH3D dataset, it achieves 0.9584 $\delta\!<\!1.25$ and 0.0803 $\mathrm{RMSE}_{\log}$, compared with 0.9729 and 0.0595 for VGGT. On ScanNet++ dataset, it improves both metrics, reaching 0.9608 and 0.1016 versus 0.9479 and 0.1106 for VGGT.

Averaged over the two datasets, DarkVGGT differs from VGGT by only 0.0008 in $\delta\!<\!1.25$ and 0.0059 in $\mathrm{RMSE}_{\log}$, exactly matches the average $\mathrm{RRA}@30$, and shows only a 0.02 decrease in $\mathrm{RTA}@30$ from 0.92 to 0.90.
This deviation is notably smaller than the teacher gap observed for Dark3R~\citep{guo2026dark3r}, which shows a 0.0147 decrease in average $\delta\!<\!1.25$ and a 0.0078 increase in $\mathrm{RMSE}_{\log}$ relative to MASt3R~\citep{leroy2024grounding}.
The degradation is more pronounced on ETH3D dataset, where $\delta\!<\!1.25$ decreases from 0.9413 to 0.8992 and $\mathrm{RMSE}_{\log}$ increases from 0.1496 to 0.1739.
For pose estimation, Dark3R likewise underperforms MASt3R by 0.025 in $\mathrm{RRA}@30$ and 0.02 in $\mathrm{RTA}@30$, while DarkVGGT achieves higher pose accuracy than Dark3R in our experiments on well-lit sequences.
Please refer to Appendix~\ref{app:rgbt_wellit_depth} for the results obtained when using RGB-T inputs on daylight sequences.

\begin{table}[t]
\centering
\small
\caption{Effectiveness of each proposed module and computational cost for RGB-T multimodal nighttime 3D geometry estimation.}
\vspace{0.5em}
\label{tab:ablation}
\setlength{\tabcolsep}{4pt}
\resizebox{\textwidth}{!}{%
\begin{tabular}{lcccccc|cc}
\toprule
 & \multicolumn{3}{c}{Depth} & \multicolumn{3}{c|}{Camera Pose} & \multicolumn{2}{c}{Cost} \\
\cmidrule(lr){2-4}\cmidrule(lr){5-7}\cmidrule(lr){8-9}
Method & AbsRel $\downarrow$ & $\delta\!<\!1.25\uparrow$ & $\text{RMSE}_{log}$ $\downarrow$ & $\text{RRA}@30\uparrow$ & $\text{RTA}@30\uparrow$ & $\text{AUC}@30\uparrow$ & \#Param. (M) & FPS \\
\midrule
AA
    & 0.1672 & 0.7512 & 0.2569
    & 0.95 & 0.77 & 61.5
    & 604.73 & 9.0 \\
LoRA
    & 0.1658 & 0.7590 & 0.2542
    & 0.96 & \thirdcell{0.79} & 62.2
    & 50.33 & 8.9 \\
LoRA + \textsc{CTFF}
    & \secondcell{0.1422} & \secondcell{0.8088} & \secondcell{0.2298}
    & \bestcell{1.00} & \secondcell{0.86} & \bestcell{68.9}
    & 107.21 & 7.5 \\
LoRA + \textsc{GSTR}
    & \thirdcell{0.1538} & \thirdcell{0.7879} & \thirdcell{0.2386}
    & \thirdcell{0.97} & \thirdcell{0.79} & \thirdcell{62.4}
    & 70.81 & 7.5 \\
\textbf{LoRA + \textsc{CTFF} + \textsc{GSTR} (Full)}
    & \bestcell{0.1373} & \bestcell{0.8243}
    & \bestcell{0.2232} & \secondcell{0.99} & \bestcell{0.89} & \secondcell{66.3}
    & 124.34 & 7.4 \\
\bottomrule
\end{tabular}
}
\end{table}

\begin{table}[t]
\centering
\small
\caption{Ablation study on the effectiveness of each loss function for both depth and pose estimation.}
\vspace{0.5em}
\label{tab:ablation2}
\setlength{\tabcolsep}{8pt}
\resizebox{0.85\textwidth}{!}{%
\begin{tabular}{lcccccc}
\toprule
 & \multicolumn{3}{c}{Depth} & \multicolumn{3}{c}{Camera Pose} \\
\cmidrule(lr){2-4}\cmidrule(lr){5-7}
Configuration
 & AbsRel $\downarrow$
 & $\delta\!<\!1.25\uparrow$
 & $\mathrm{RMSE}_{\log}\downarrow$
 & $\mathrm{RRA}@5 \uparrow$
 & $\mathrm{RTA}@5 \uparrow$
 & $\mathrm{AUC}@30 \uparrow$ \\
\midrule
\textbf{Full}
 & \bestcell{0.1373} & \bestcell{0.8243} & \bestcell{0.2232}
 & \bestcell{0.68} & \bestcell{0.45} & \secondcell{66.3} \\
w/o $\mathcal{L}_{\mathrm{sparse}}$
 & \secondcell{0.1381} & \secondcell{0.8188} & \secondcell{0.2249}
 & \bestcell{0.68} & \bestcell{0.45} & \bestcell{69.2} \\
w/o $\mathcal{L}_{\mathrm{recon}}$
 & \thirdcell{0.1404} & \thirdcell{0.8151} & \thirdcell{0.2274}
 & \bestcell{0.68} & \thirdcell{0.44} & \thirdcell{65.9} \\
w/o $\mathcal{L}_{\mathrm{edge}}$
 & 0.1418 & 0.8145 & 0.2287
 & 0.67 & 0.43 & 65.3 \\
\bottomrule
\end{tabular}%
}
\end{table}

\paragraph{Ablation study.}
Table~\ref{tab:ablation} demonstrates the effectiveness of each proposed module for RGB-T nighttime 3D geometry estimation. AA denotes full fine-tuning of the alternating attention blocks without LoRA, while \textsc{CTFF} and \textsc{GSTR} denote the complementary thermal feature factorization module and geometry-shared thermal routing, respectively.
Adding LoRA to the AA baseline yields only modest gains, improving AbsRel from 0.1672 to 0.1658, $\delta\!<\!1.25$ from 0.7512 to 0.7590, and $\mathrm{AUC}@30$ from 61.5 to 62.2.
In contrast, \textsc{CTFF} provides the largest standalone improvement, reducing AbsRel to 0.1422 and $\mathrm{RMSE}_{\log}$ to 0.2298, while also increasing $\mathrm{RRA}@30$, $\mathrm{RTA}@30$, and $\mathrm{AUC}@30$ to 1.00, 0.86, and 68.9, respectively.
\textsc{GSTR} provides a smaller standalone gain, but complements \textsc{CTFF} in the full model, reaching 0.1373 AbsRel, 0.8243 $\delta\!<\!1.25$, and 0.89 $\mathrm{RTA}@30$ (Table~\ref{tab:ablation}). Relative to \textsc{CTFF} alone, these depth and translation gains are accompanied by decreases in $\mathrm{RRA}@30$ from 1.00 to 0.99 and aggregate $\mathrm{AUC}@30$ from 68.9 to 66.3.
Selected sequence-level comparisons between LoRA+CTFF and the full model are provided in Appendix~\ref{app:ctff_full}.
Compared with AA, the full model improves all metrics, including a 17.9\% reduction in AbsRel and a 7.8\% increase in $\mathrm{AUC}@30$.
Regarding computational efficiency, our full model achieves superior performance with the proposed modules while requiring 79.4\% fewer learnable parameters than full AA fine-tuning, reducing the parameter count from 604.73M to 124.34M and achieving an inference speed of 7.4 FPS. These results highlight the benefit of combining thermal feature modulation with shared-subspace routing for nighttime geometry estimation.

Table~\ref{tab:ablation2} ablates the three regularizers by removing each term individually. Since pose differences are relatively small, we report $\mathrm{RRA}@5$ and $\mathrm{RTA}@5$ for a more fine-grained comparison. Dropping $\mathcal{L}_{\mathrm{edge}}$ causes the largest AbsRel increase, from 0.1373 to 0.1418, as it supervises residual allocation using thermal--depth edge disagreement. Without $\mathcal{L}_{\mathrm{recon}}$, AbsRel degrades to 0.1404, as the thermal reconstruction constraint on the private subspace is removed. $\mathcal{L}_{\mathrm{sparse}}$ has the smallest AbsRel impact, but its removal lowers $\delta\!<\!1.25$, suggesting that limiting residual allocation benefits depth estimation. Although this variant improves $\mathrm{AUC}@30$ from 66.3 to 69.2, it does so at the cost of degraded depth estimation. These findings support the complementary roles of allocation regularization and private-subspace reconstruction.

\section{Conclusion}
\label{sec:conclusion}
\vspace{-3mm}
We presented DarkVGGT, a structured RGB-T adaptation of pretrained VGGT that leverages thermal observations for robust 3D geometry estimation in low-visibility scenes.
DarkVGGT modulates content-preserving and high-frequency residual thermal pathways through learned complementary allocation weights.
Additionally, geometry-shared thermal routing injects gated corrections derived from thermal--RGB feature differences in a shared latent space.
Across low-visibility RGB-T benchmarks, DarkVGGT outperforms existing RGB-only and RGB-T feed-forward models in depth estimation and in camera pose estimation averaged across the three datasets.
Experiments on well-lit datasets including ETH3D and ScanNet++ show that DarkVGGT largely preserves the original capability of VGGT, thereby reducing the daylight tax of low-light adaptation.
Ablation studies further indicate that complementary thermal feature factorization drives the main gains, while geometry-shared thermal routing provides complementary refinement with little runtime overhead.
Overall, DarkVGGT demonstrates that thermal guidance can improve dark-scene 3D perception while largely preserving RGB robustness in well-lit settings.

\paragraph{Limitation.}
Our experiments assume moderately aligned paired RGB-T inputs, which may limit applicability to weakly aligned or unpaired settings. Addressing such cases would require additional cross-modal correspondence or correction mechanisms. Cross-modal appearance differences can also weaken correspondence when ordinary glass is transparent in visible light but opaque in long-wave infrared, or when internal heat sources induce surface thermal patterns without corresponding visible texture. Our modules also introduce a few hyperparameters for residual scaling, sparsity, and fusion budgeting, which may require validation-based tuning or sensor-specific calibration. These limitations are orthogonal to the VGGT backbone.
\paragraph{Future work.}
Future work includes adaptive RGB-T fusion, weaker supervision settings such as partially aligned or unpaired RGB-T data, and extensions to other multimodal geometry cues such as event cameras or LiDAR.
While this direction may support safer robotic and autonomous perception in low-visibility environments, RGB-T systems can also raise privacy concerns in public or sensitive spaces due to thermal sensing.
Future deployment should therefore follow appropriate privacy, safety, and data-governance safeguards.

\section{Acknowledgements}
This work was supported in part by GPU hardware and cloud credits provided to the Texas A\&M University PHAI Lab through the NVIDIA Academic Grant Program. We also acknowledge computational resources provided by the Texas A\&M VISION GPU Cluster. Z. Fan was supported in part by research gifts from Meta and the US AI Glasses Impact Grants program.

\small
\bibliographystyle{plainnat}
\bibliography{neurips_2026}

@String(ICLR = {Int. Conf. Learn. Represent.})

@String(AAAI = {AAAI})

@String(ACCESS  = {IEEE access})

@String(ICRA  = {IEEE Int. Conf. Robotics and Automation})

@String(IROS  = {IEEE/RSJ Int. Conf. Intell. Robots and Systems})

@String(DV  = {Int. Conf. 3D Vis.})

@String(SENSORS = {Sensors})

@String(ICLR  = {ICLR})

@inproceedings{wang2024dust3r,
  title={{DUSt3R}: Geometric {3D} vision made easy},
  author={Wang, Shuzhe and Leroy, Vincent and Cabon, Yohann and Chidlovskii, Boris and Revaud, Jerome},
  booktitle={Proceedings of the IEEE/CVF conference on computer vision and pattern recognition},
  pages={20697--20709},
  year={2024}
}

@inproceedings{leroy2024grounding,
  title={Grounding image matching in {3D} with {MASt3R}},
  author={Leroy, Vincent and Cabon, Yohann and Revaud, J{\'e}r{\^o}me},
  booktitle={European conference on computer vision},
  pages={71--91},
  year={2024},
  organization={Springer}
}

@inproceedings{duisterhof2025mast3r,
  title={{MASt3R-SfM}: A fully-integrated solution for unconstrained structure-from-motion},
  author={Duisterhof, Bardienus Pieter and Zust, Lojze and Weinzaepfel, Philippe and Leroy, Vincent and Cabon, Yohann and Revaud, Jerome},
  booktitle={2025 International Conference on 3D Vision (3DV)},
  pages={1--10},
  year={2025},
  organization={IEEE}
}

@inproceedings{schonberger2016structure,
  title={Structure-from-motion revisited},
  author={Schonberger, Johannes L and Frahm, Jan-Michael},
  booktitle={Proceedings of the IEEE conference on computer vision and pattern recognition},
  pages={4104--4113},
  year={2016}
}

@article{skorokhodov2026sear,
  title={{SEAR}: Simple and Efficient Adaptation of Visual Geometric Transformers for {RGB+Thermal} {3D} Reconstruction},
  author={Skorokhodov, Vsevolod and Xu, Chenghao and Sun, Shuo and Fink, Olga and Mielle, Malcolm},
  journal={arXiv preprint arXiv:2603.18774},
  year={2026}
}

@article{brenner2023rgb,
  title={{RGB-D} and thermal sensor fusion: A systematic literature review},
  author={Brenner, Martin and Reyes, Napoleon H and Susnjak, Teo and Barczak, Andre LC},
  journal={{IEEE} Access},
  volume={11},
  pages={82410--82442},
  year={2023},
  publisher={IEEE}
}

@article{tan2026masked,
  title={Masked Depth Modeling for Spatial Perception},
  author={Tan, Bin and Sun, Changjiang and Qin, Xiage and Adai, Hanat and Fu, Zelin and Zhou, Tianxiang and Zhang, Han and Xu, Yinghao and Zhu, Xing and Shen, Yujun and others},
  journal={arXiv preprint arXiv:2601.17895},
  year={2026}
}

@inproceedings{zhang2025flare,
  title={{FLARE}: Feed-forward geometry, appearance and camera estimation from uncalibrated sparse views},
  author={Zhang, Shangzhan and Wang, Jianyuan and Xu, Yinghao and Xue, Nan and Rupprecht, Christian and Zhou, Xiaowei and Shen, Yujun and Wetzstein, Gordon},
  booktitle={Proceedings of the Computer Vision and Pattern Recognition Conference},
  pages={21936--21947},
  year={2025}
}

@article{usamentiaga2014infrared,
  title={Infrared thermography for temperature measurement and non-destructive testing},
  author={Usamentiaga, Rub{\'e}n and Venegas, Pablo and Guerediaga, Jon and Vega, Laura and Molleda, Julio and Bulnes, Francisco G},
  journal={Sensors},
  volume={14},
  number={7},
  pages={12305--12348},
  year={2014},
  publisher={MDPI}
}

@article{maheshwari2026anythermal,
  title={{AnyThermal}: Towards Learning Universal Representations for Thermal Perception},
  author={Maheshwari, Parv and Karhade, Jay and Chawla, Yogesh and Adu, Isaiah and Heisen, Florian and Porco, Andrew and Jong, Andrew and Liu, Yifei and Pitla, Santosh and Scherer, Sebastian and others},
  journal={arXiv preprint arXiv:2602.06203},
  year={2026}
}

@article{kweon2025mrgs,
  title={{MrGS}: Multi-modal Radiance Fields with {3D} Gaussian Splatting for {RGB-Thermal} Novel View Synthesis},
  author={Kweon, Minseong and Kim, Janghyun and Shin, Ukcheol and Park, Jinsun},
  journal={arXiv preprint arXiv:2511.22997},
  year={2025}
}

@inproceedings{he2023darkfeat,
  title={{DarkFeat}: Noise-robust feature detector and descriptor for extremely low-light {RAW} images},
  author={He, Yuze and Hu, Yubin and Zhao, Wang and Li, Jisheng and Liu, Yong-Jin and Han, Yuxing and Wen, Jiangtao},
  booktitle={Proceedings of the AAAI conference on artificial intelligence},
  volume={37},
  pages={826--834},
  year={2023}
}

@inproceedings{schonberger2016pixelwise,
  title={Pixelwise view selection for unstructured multi-view stereo},
  author={Sch{\"o}nberger, Johannes L and Zheng, Enliang and Frahm, Jan-Michael and Pollefeys, Marc},
  booktitle={European conference on computer vision},
  pages={501--518},
  year={2016},
  organization={Springer}
}

@article{selvaratnam20253d,
  title={{3D} reconstruction in robotics: A comprehensive review},
  author={Selvaratnam, Dharmendra and Bazazian, Dena},
  journal={Computers \& Graphics},
  volume={130},
  pages={104256},
  year={2025},
  publisher={Elsevier}
}

@article{hu2022lora,
  title={{LoRA}: Low-rank adaptation of large language models},
  author={Hu, Edward J and Shen, Yelong and Wallis, Phillip and Allen-Zhu, Zeyuan and Li, Yuanzhi and Wang, Shean and Wang, Liang and Chen, Weizhu and others},
  journal={Iclr},
  volume={1},
  number={2},
  pages={3},
  year={2022}
}

@article{shin2021self,
  title   = {Self-Supervised Depth and Ego-Motion Estimation for Monocular Thermal Video Using Multi-Spectral Consistency Loss},
  author  = {Shin, Ukcheol and Lee, Kyunghyun and Lee, Seokju and Kweon, In So},
  journal = {IEEE Robotics and Automation Letters},
  volume  = {7},
  number  = {2},
  pages   = {1103--1110},
  year    = {2021}
}

@inproceedings{shin2023deep,
  title     = {Deep Depth Estimation From Thermal Image},
  author    = {Shin, Ukcheol and Park, Jinsun and Kweon, In So},
  booktitle = {Proceedings of the IEEE/CVF Conference on Computer Vision and Pattern Recognition},
  year      = {2023}
}

@article{guo2022cross,
  title   = {Unsupervised Visible-light Images Guided Cross-Spectrum Depth Estimation from Dual-Modality Cameras},
  author  = {Guo, Yubin and Jiang, Haobo and Qi, Xinlei and Xie, Jin and Xu, Cheng-Zhong and Kong, Hui},
  journal = {arXiv preprint arXiv:2205.00257},
  year    = {2022}
}

@inproceedings{kwon2024misaligned,
  title     = {Multi-modal Depth Estimation from Misaligned Thermal and {RGB} Images},
  author    = {Kwon, Byeongjun and Kim, Munchurl},
  booktitle = {Proceedings of the Korean Institute of Broadcast and Media Engineers Summer Conference},
  pages     = {912--915},
  year      = {2024}
}

@article{zuo2025monother,
  title   = {{MonoTher-Depth}: Enhancing Thermal Depth Estimation via Confidence-Aware Distillation},
  author  = {Zuo, Xingxing and Ranganathan, Nikhil and Lee, Connor and Gkioxari, Georgia and Chung, Soon-Jo},
  journal = {IEEE Robotics and Automation Letters},
  volume  = {10},
  number  = {3},
  pages   = {2830--2837},
  year    = {2025}
}

@inproceedings{shin2025bridging,
  title={Bridging spectral-wise and multi-spectral depth estimation via geometry-guided contrastive learning},
  author={Shin, Ukcheol and Lee, Kyunghyun and Oh, Jean},
  booktitle={2025 IEEE International Conference on Robotics and Automation (ICRA)},
  pages={6299--6305},
  year={2025},
  organization={IEEE}
}

@article{hassan2024thermonerf,
  title   = {{ThermoNeRF}: Joint {RGB} and Thermal Novel View Synthesis for Building Facades using Multimodal Neural Radiance Fields},
  author  = {Hassan, Mariam and Forest, Florent and Fink, Olga and Mielle, Malcolm},
  journal = {arXiv preprint arXiv:2403.12154},
  year    = {2024}
}

@inproceedings{lin2024thermalnerf,
  title={Thermalnerf: Thermal radiance fields},
  author={Lin, Yvette Y and Pan, Xin-Yi and Fridovich-Keil, Sara and Wetzstein, Gordon},
  booktitle={2024 IEEE International Conference on Computational Photography (ICCP)},
  pages={1--12},
  year={2024},
  organization={IEEE}
}

@inproceedings{chen2024thermal3dgs,
  title     = {{Thermal3D-GS}: Physics-induced {3D} Gaussians for Thermal Infrared Novel-view Synthesis},
  author    = {Chen, Qian and Shu, Shihao and Bai, Xiangzhi},
  booktitle = {European Conference on Computer Vision},
  year      = {2024}
}

@article{zhang2026multimodal,
  title={Multimodal fusion on low-quality data: A comprehensive survey},
  author={Zhang, Qingyang and Wei, Yake and Han, Zongbo and Fu, Huazhu and Peng, Xi and Hu, Qinghua and Deng, Cheng and Xu, Cai and Wen, Jie and Hu, Di and others},
  journal={Information Fusion},
  pages={104437},
  year={2026},
  publisher={Elsevier}
}

@article{reza2004realization,
  title={Realization of the contrast limited adaptive histogram equalization (CLAHE) for real-time image enhancement},
  author={Reza, Ali M},
  journal={Journal of VLSI signal processing systems for signal, image and video technology},
  volume={38},
  number={1},
  pages={35--44},
  year={2004},
  publisher={Springer}
}

@article{ghari2024pedestrian,
  title={Pedestrian detection in low-light conditions: A comprehensive survey},
  author={Ghari, Bahareh and Tourani, Ali and Shahbahrami, Asadollah and Gaydadjiev, Georgi},
  journal={Image and Vision Computing},
  volume={148},
  pages={105106},
  year={2024},
  publisher={Elsevier}
}

@article{arnold2019survey,
  title={A survey on 3d object detection methods for autonomous driving applications},
  author={Arnold, Eduardo and Al-Jarrah, Omar Y and Dianati, Mehrdad and Fallah, Saber and Oxtoby, David and Mouzakitis, Alex},
  journal={IEEE Transactions on Intelligent Transportation Systems},
  volume={20},
  number={10},
  pages={3782--3795},
  year={2019},
  publisher={IEEE}
}

@article{cho2025vr,
  title={VR-Drive: Viewpoint-Robust End-to-End Driving with Feed-Forward 3D Gaussian Splatting},
  author={Cho, Hoonhee and Kang, Jae-Young and Lee, Giwon and Yang, Hyemin and Park, Heejun and Jung, Seokwoo and Yoon, Kuk-Jin},
  journal={arXiv preprint arXiv:2510.23205},
  year={2025}
}

@article{yang2026robo3r,
  title={Robo3R: Enhancing Robotic Manipulation with Accurate Feed-Forward 3D Reconstruction},
  author={Yang, Sizhe and Xu, Linning and Li, Hao and Mu, Juncheng and Zeng, Jia and Lin, Dahua and Pang, Jiangmiao},
  journal={arXiv preprint arXiv:2602.10101},
  year={2026}
}

@article{lu2024drivingrecon,
  title={DrivingRecon: Large 4D Gaussian reconstruction model for autonomous driving},
  author={Lu, Hao and Xu, Tianshuo and Zheng, Wenzhao and Zhang, Yunpeng and Zhan, Wei and Du, Dalong and Tomizuka, Masayoshi and Keutzer, Kurt and Chen, Yingcong},
  journal={arXiv preprint arXiv:2412.09043},
  year={2024}
}

@article{han2025d,
  title={D\^{} 2USt3R: Enhancing 3D Reconstruction with 4D Pointmaps for Dynamic Scenes},
  author={Han, Jisang and An, Honggyu and Jung, Jaewoo and Narihira, Takuya and Seo, Junyoung and Fukuda, Kazumi and Kim, Chaehyun and Hong, Sunghwan and Mitsufuji, Yuki and Kim, Seungryong},
  journal={arXiv e-prints},
  pages={arXiv--2504},
  year={2025}
}

@article{maggio2025vggt,
  title={Vggt-slam: Dense rgb slam optimized on the sl (4) manifold},
  author={Maggio, Dominic and Lim, Hyungtae and Carlone, Luca},
  journal={arXiv preprint arXiv:2505.12549},
  year={2025}
}

@inproceedings{murai2025mast3r,
  title={Mast3r-slam: Real-time dense slam with 3d reconstruction priors},
  author={Murai, Riku and Dexheimer, Eric and Davison, Andrew J},
  booktitle={Proceedings of the Computer Vision and Pattern Recognition Conference},
  pages={16695--16705},
  year={2025}
}

@inproceedings{yang2025fast3r,
  title={Fast3r: Towards 3d reconstruction of 1000+ images in one forward pass},
  author={Yang, Jianing and Sax, Alexander and Liang, Kevin J and Henaff, Mikael and Tang, Hao and Cao, Ang and Chai, Joyce and Meier, Franziska and Feiszli, Matt},
  booktitle={Proceedings of the Computer Vision and Pattern Recognition Conference},
  pages={21924--21935},
  year={2025}
}

@article{fang2026more,
  title={MoRe: Motion-aware Feed-forward 4D Reconstruction Transformer},
  author={Fang, Juntong and Chen, Zequn and Zhang, Weiqi and Di, Donglin and Zhang, Xuancheng and Yang, Chengmin and Liu, Yu-Shen},
  journal={arXiv preprint arXiv:2603.05078},
  year={2026}
}

@article{zhang2024monst3r,
  title={{MonST3R}: A simple approach for estimating geometry in the presence of motion},
  author={Zhang, Junyi and Herrmann, Charles and Hur, Junhwa and Jampani, Varun and Darrell, Trevor and Cole, Forrester and Sun, Deqing and Yang, Ming-Hsuan},
  journal={arXiv preprint arXiv:2410.03825},
  year={2024}
}

@inproceedings{wang20253d,
  title={{3D} reconstruction with spatial memory},
  author={Wang, Hengyi and Agapito, Lourdes},
  booktitle={2025 International Conference on 3D Vision (3DV)},
  pages={78--89},
  year={2025},
  organization={IEEE}
}

@inproceedings{cabon2025must3r,
  title={{MUSt3R}: Multi-view network for stereo {3D} reconstruction},
  author={Cabon, Yohann and Stoffl, Lucas and Antsfeld, Leonid and Csurka, Gabriela and Chidlovskii, Boris and Revaud, Jerome and Leroy, Vincent},
  booktitle={Proceedings of the IEEE/CVF Conference on Computer Vision and Pattern Recognition},
  pages={1050--1060},
  year={2025}
}

@inproceedings{wang2025vggt,
  title={{VGGT}: Visual Geometry Grounded Transformer},
  author={Wang, Jianyuan and Chen, Minghao and Karaev, Nikita and Vedaldi, Andrea and Rupprecht, Christian and Novotny, David},
  booktitle={Proceedings of the Computer Vision and Pattern Recognition Conference},
  pages={5294--5306},
  year={2025}
}

@article{oquab2023dinov2,
  title={{DINOv2}: Learning robust visual features without supervision},
  author={Oquab, Maxime and Darcet, Timoth{\'e}e and Moutakanni, Th{\'e}o and Vo, Huy and Szafraniec, Marc and Khalidov, Vasil and Fernandez, Pierre and Haziza, Daniel and Massa, Francisco and El-Nouby, Alaaeldin and others},
  journal={arXiv preprint arXiv:2304.07193},
  year={2023}
}

@article{lee2022vivid++,
  title={{ViViD++}: Vision for visibility dataset},
  author={Lee, Alex Junho and Cho, Younggun and Shin, Young-sik and Kim, Ayoung and Myung, Hyun},
  journal={IEEE Robotics and Automation Letters},
  volume={7},
  number={3},
  pages={6282--6289},
  year={2022},
  publisher={IEEE}
}

@article{wu2025eag3r,
  title={{EAG3R}: Event-Augmented {3D} Geometry Estimation for Dynamic and Extreme-Lighting Scenes},
  author={Wu, Xiaoshan and Yu, Yifei and Lyu, Xiaoyang and Huang, Yihua and Wang, Bo and Zhang, Baoheng and Wang, Zhongrui and Qi, Xiaojuan},
  journal={arXiv preprint arXiv:2512.00771},
  year={2025}
}

@article{ren2026eventvggt,
  title={{EventVGGT}: Exploring Cross-Modal Distillation for Consistent Event-based Depth Estimation},
  author={Ren, Yinrui and Zhu, Jinjing and Chen, Kanghao and Li, Zhuoxiao and Ou, Jing and Cao, Zidong and Hua, Tongyan and Shi, Peilun and Fu, Yingchun and Zhao, Wufan and others},
  journal={arXiv preprint arXiv:2603.09385},
  year={2026}
}

@article{su2024roformer,
  title={{RoFormer}: Enhanced transformer with rotary position embedding},
  author={Su, Jianlin and Ahmed, Murtadha and Lu, Yu and Pan, Shengfeng and Bo, Wen and Liu, Yunfeng},
  journal={Neurocomputing},
  volume={568},
  pages={127063},
  year={2024},
  publisher={Elsevier}
}

@book{Vollmer_2010,
  title     = {Infrared Thermal Imaging: Fundamentals, Research and Applications},
  author    = {Vollmer, Michael and M{\"o}llmann, Klaus-Peter},
  publisher = {Wiley},
  year      = {2010},
  month     = aug,
  isbn      = {9783527630868},
  doi       = {10.1002/9783527630868},
  url       = {http://dx.doi.org/10.1002/9783527630868}
}

@inproceedings{narayanan2026dual,
  title = {{Dual Band Thermal Videography: Separating Time-Varying Reflection and Emission Near Ambient Conditions}},
  author={Narayanan, Sriram and Ramanagopal, Mani and Narasimhan, Srinivasa},
  booktitle={Proceedings of the IEEE/CVF Conference on Computer Vision and Pattern Recognition},
  pages={199--208},
  year={2026}
}

@article{lu2024thermalgaussian,
  title={{ThermalGaussian}: Thermal {3D} Gaussian Splatting},
  author={Lu, Rongfeng and Chen, Hangyu and Zhu, Zunjie and Qin, Yuhang and Lu, Ming and Zhang, Le and Yan, Chenggang and Xue, Anke},
  journal={arXiv preprint arXiv:2409.07200},
  year={2024}
}

@article{guo2026dark3r,
  title={{Dark3R}: Learning Structure from Motion in the Dark},
  author={Guo, Andrew Y and Malik, Anagh and Tedla, SaiKiran and Dai, Yutong and Qin, Yiqian and Salehe, Zach and Attal, Benjamin and Nousias, Sotiris and Kutulakos, Kyros and Lindell, David B},
  journal={arXiv preprint arXiv:2603.05330},
  year={2026}
}

@inproceedings{yun2022sthereo,
  title={{STheReO}: Stereo thermal dataset for research in odometry and mapping},
  author={Yun, Seungsang and Jung, Minwoo and Kim, Jeongyun and Jung, Sangwoo and Cho, Younghun and Jeon, Myung-Hwan and Kim, Giseop and Kim, Ayoung},
  booktitle={2022 IEEE/RSJ International Conference on Intelligent Robots and Systems (IROS)},
  pages={3857--3864},
  year={2022},
  organization={IEEE}
}

@article{depthanything3,
  title={{Depth Anything 3}: Recovering the visual space from any views},
  author={Haotong Lin and Sili Chen and Jun Hao Liew and Donny Y. Chen and Zhenyu Li and Guang Shi and Jiashi Feng and Bingyi Kang},
  journal={arXiv preprint arXiv:2511.10647},
  year={2025}
}

@inproceedings{ranftl2021vision,
  title={Vision transformers for dense prediction},
  author={Ranftl, Ren{\'e} and Bochkovskiy, Alexey and Koltun, Vladlen},
  booktitle={Proceedings of the IEEE/CVF international conference on computer vision},
  pages={12179--12188},
  year={2021}
}

@article{xiao2025thermalgen,
  title={{ThermalGen}: Style-Disentangled Flow-Based Generative Models for {RGB-to-Thermal} Image Translation},
  author={Xiao, Jiuhong and Nayak, Roshan and Zhang, Ning and Tortei, Daniel and Loianno, Giuseppe},
  journal={arXiv preprint arXiv:2509.24878},
  year={2025}
}

@inproceedings{schops2017multi,
  title={A multi-view stereo benchmark with high-resolution images and multi-camera videos},
  author={Schops, Thomas and Schonberger, Johannes L and Galliani, Silvano and Sattler, Torsten and Schindler, Konrad and Pollefeys, Marc and Geiger, Andreas},
  booktitle={Proceedings of the IEEE conference on computer vision and pattern recognition},
  pages={3260--3269},
  year={2017}
}

@inproceedings{yeshwanth2023scannet++,
  title={{ScanNet++}: A high-fidelity dataset of {3D} indoor scenes},
  author={Yeshwanth, Chandan and Liu, Yueh-Cheng and Nie{\ss}ner, Matthias and Dai, Angela},
  booktitle={Proceedings of the IEEE/CVF International Conference on Computer Vision},
  pages={12--22},
  year={2023}
}

@inproceedings{wu2021text,
  title={A text-centered shared-private framework via cross-modal prediction for multimodal sentiment analysis},
  author={Wu, Yang and Lin, Zijie and Zhao, Yanyan and Qin, Bing and Zhu, Li-Nan},
  booktitle={Findings of the association for computational linguistics: ACL-IJCNLP 2021},
  pages={4730--4738},
  year={2021}
}

\normalsize
\appendix

\section{Methodological details}

\subsection{Architecture and adaptation}
\label{sec:supp-method}

\paragraph{LoRA and camera tokens.}
After loading the pretrained VGGT~\citep{wang2025vggt} checkpoint, we initialize the thermal camera token by duplicating the pretrained camera token, following the same token-initialization strategy as SEAR~\citep{skorokhodov2026sear}. This allows both modalities to start from the same geometric token prior. We then insert rank-64 LoRA~\citep{hu2022lora} adapters, shared across the RGB and thermal streams, into the \texttt{qkv}, \texttt{proj}, \texttt{fc1}, and \texttt{fc2} projections of both the frame-attention and global-attention blocks, while keeping the original backbone weights frozen. The parameter counts in Table~\ref{tab:ablation} refer to learnable parameters. The VGGT aggregator contains 24 alternating-attention rounds, each comprising one frame-attention and one global-attention block, for 48 blocks in total. The input-plus-output dimensions of the adapted projections sum to 16,384 per block, yielding $48\times64\times16{,}384=50{,}331{,}648\approx50.33$M learnable LoRA parameters.

\paragraph{Late geometry-shared thermal routing.}
The geometry-shared decomposition uses ranks $(d_s, d_p, d_{\mathrm{mi}})=(96,192,128)$, with the distillation projection heads in Eq.~\ref{eq:l-distill} implemented as single linear maps $\mathbb{R}^{d_s}\!\to\!\mathbb{R}^{d_{\mathrm{mi}}}$. The pathway is instantiated in the last $k=8$ AA blocks, with the residual scale initialized to $\alpha_{\mathrm{gstr}}=\sigma(-2.2)\approx 0.1$ and the corrective injection and gating heads zero-initialized, so the branch is effectively inactive at initialization and is activated only as training learns a consistent corrective signal. The reconstruction and alignment losses $\mathcal{L}_{\mathrm{recon}}$ and $\mathcal{L}_{\mathrm{distill}}$ follow a cosine warmup from $0.01$ to $0.10$ over the first $1500$ of $4000$ gradient-accumulation steps. Without this warmup, applying the target weight from step zero pulls the private subspace prematurely toward thermal reconstruction and degrades early depth/pose accuracy.

\paragraph{Thermal positional adaptation.}
To account for residual spatial offsets between RGB and thermal views, we introduce a learnable 2D thermal RoPE~\citep{su2024roformer} offset. Thermal tokens share the rotary positional grid of RGB tokens, with an additive offset $(\Delta y,\Delta x)$ applied before rotary encoding. The offset is initialized to zero, allowing the multimodal model to start from the aligned RGB prior and learn cross-sensor positional corrections only when supported by the training signal. This lightweight mechanism preserves the RGB branch while adapting only the thermal positional coordinates.

\subsection{Geometry-Shared Thermal Routing visualization}

\begin{figure}[!htb]
    \centering
    \small
    \setlength{\tabcolsep}{2pt}
    \renewcommand{\arraystretch}{1.0}

    \begin{tabular}{@{}ccc@{}}
        \text{RGB} & \text{Thermal} & \text{Injection} \\[-0.2em]

        \includegraphics[width=0.28\textwidth]{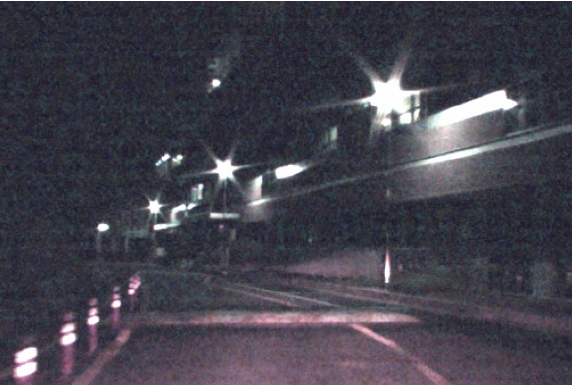}
        & \includegraphics[width=0.28\textwidth]{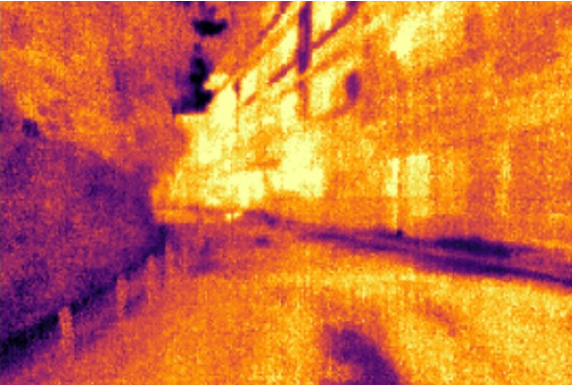}
        & \includegraphics[width=0.28\textwidth]{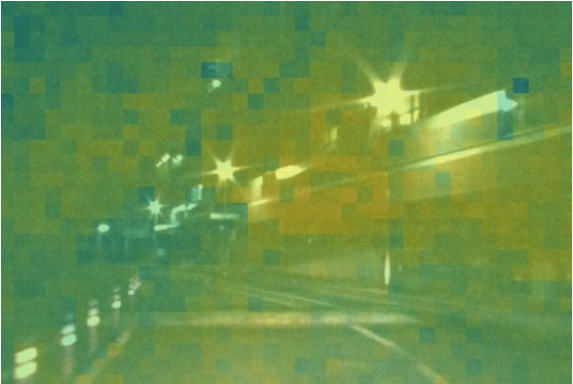} \\[0.6em]

        \includegraphics[width=0.28\textwidth]{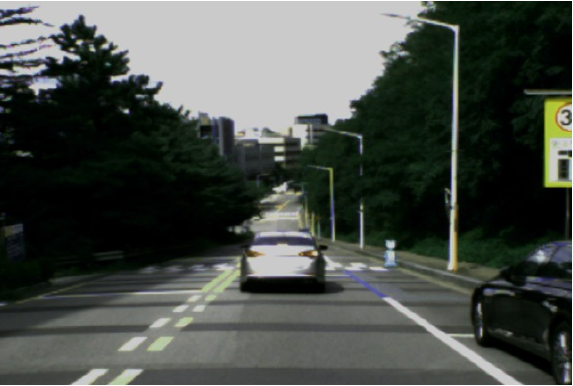}
        & \includegraphics[width=0.28\textwidth]{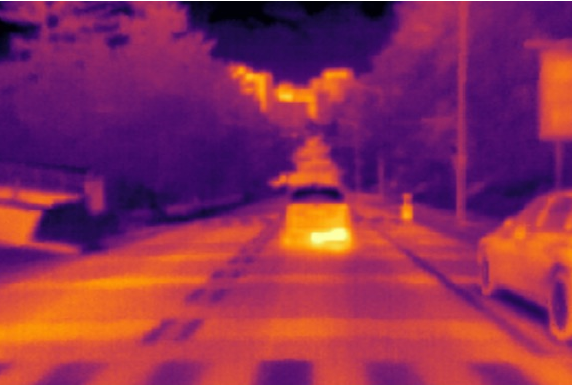}
        & \includegraphics[width=0.28\textwidth]{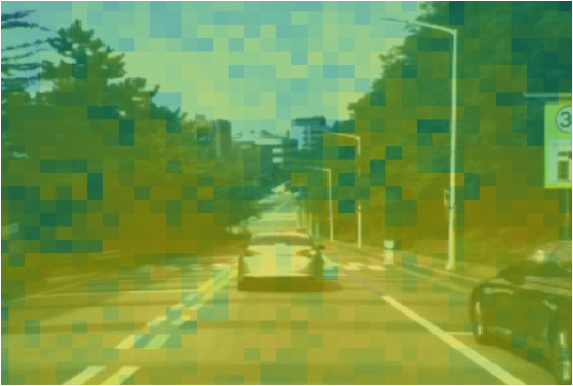} \\
    \end{tabular}

    \caption{Reliability-gated injection samples during training in dark and light scenes.}
    \label{fig:sp_injection_grid}
\end{figure}

Figure~\ref{fig:sp_injection_grid} visualizes the per-token $\ell_2$ norm of the gated geometry-shared token injection $\boldsymbol{g}^{(\ell)}_{\mathrm{s}} \odot W_{\mathrm{up}}\!\left(\sg[\bar{\mathbf{u}}^{\mathrm{s},(\ell)}]-\bar{\mathbf{v}}^{\mathrm{s},(\ell)}\right)$ from Eq.~\ref{eq:gstr-injection}, evaluated on paired RGB-T inputs sampled from day and night training scenes. In the dark night scene shown in the top row, the RGB image is severely underexposed and dominated by saturated headlights, while the corresponding thermal frame depicts the road geometry, surrounding buildings, and structural edges clearly. The geometry-shared routing magnitude in this case concentrates on the unlit road surface and the building facades on either side, which are precisely the regions where the RGB pathway carries little usable information and the thermal stream is informative. In the well-lit daytime scene shown in the bottom row, where the RGB image is properly exposed and rich in texture, the injection magnitude remains uniformly low across the entire field of view, with no regions of pronounced activation. At the deepest active AA block, the per-token feature-conditioned gate $\boldsymbol{g}^{(\ell)}_{\mathrm{s}}\in[0,1]^P$ modulates the shared-thermal residual before RGB-token injection. This contrast suggests that the gate suppresses geometry-shared residual injection when RGB cues are sufficient and selectively routes thermal-recovered structure where RGB fails.

\subsection{Edge supervision and target sensitivity}
The depth edge map used by $\mathcal{L}_{\mathrm{edge}}$ is computed from ground-truth depth using Sobel filters. Thermal and depth edge magnitudes are pooled over each patch to construct the soft target $a_{\mathrm{h,tar}}^{(\ell)}$. This supervision is used only during training.

\paragraph{Sensitivity of the edge target.}
The edge-disagreement target uses sharpness $\beta=8$ and margin $\tau=0.15$. For positive $\beta$, changing sharpness preserves the ordering of the disagreement scores and the target's $0.5$ decision boundary, while changing target confidence. In the sensitivity evaluation, the measured Spearman rank correlation is at least 0.999. Halving sharpness to $\beta=4$ increases target entropy by approximately $1.8\times$. Doubling the margin to $\tau=0.30$ changes the fraction of flagged patches by at most 8.5 percentage points. Retraining with these changes individually alters AbsRel by at most 0.0015 on each benchmark, with aggregate AbsRel changes of $-0.0005$ and $-0.0008$, respectively. These small changes support robustness to the tested sharpness and margin settings.

\section{Additional experiments and analysis}

\subsection{DarkVGGT vs. SEAR}
\label{app:darkvggt_vs_sear}
\begin{figure*}[!t]
    \centering
    \setlength{\tabcolsep}{1.5pt}
    \renewcommand{\arraystretch}{1.0}

    \begin{tabular}{@{}cccc@{}}
        \multicolumn{2}{c}{\text{SEAR}} & \multicolumn{2}{c}{\textbf{DarkVGGT (Ours)}} \\
        \cmidrule(lr){1-2}\cmidrule(lr){3-4}
        \text{RGB} & \text{Thermal} & \text{RGB} & \text{Thermal} \\
        \midrule
        \includegraphics[width=0.23\textwidth]{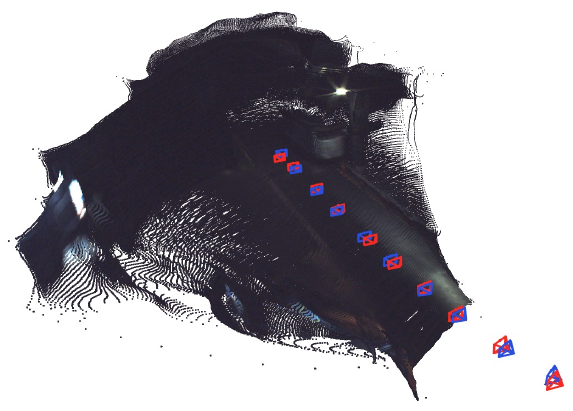} &
        \includegraphics[width=0.23\textwidth]{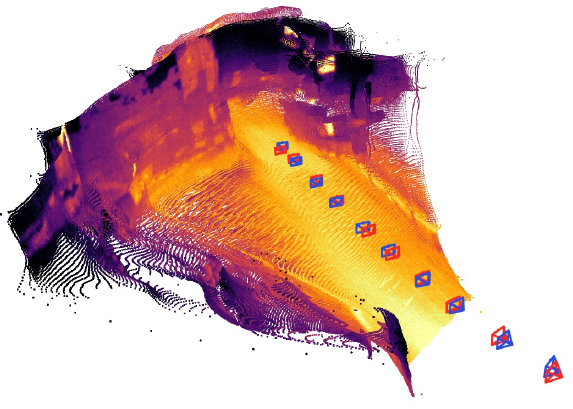} &
        \includegraphics[width=0.23\textwidth]{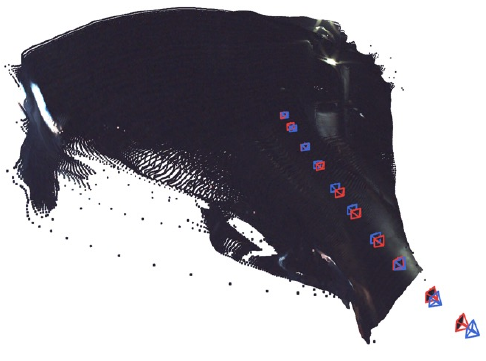} &
        \includegraphics[width=0.23\textwidth]{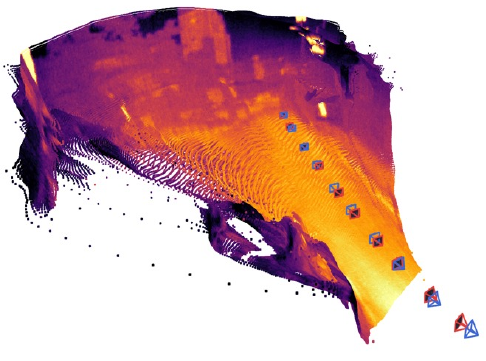} \\

        \includegraphics[width=0.23\textwidth]{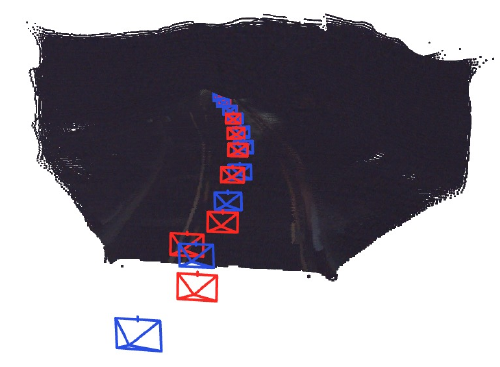} &
        \includegraphics[width=0.23\textwidth]{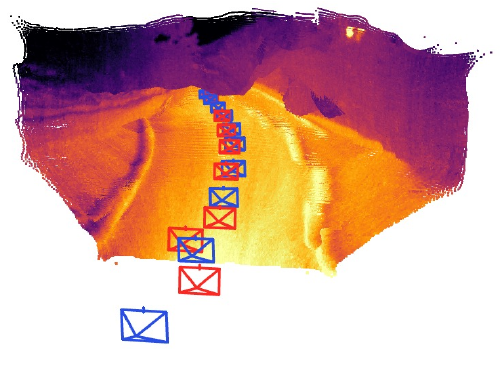} &
        \includegraphics[width=0.23\textwidth]{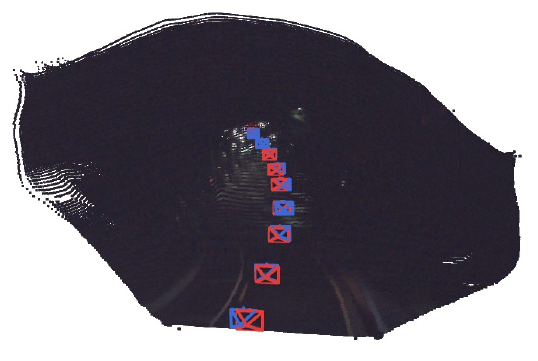} &
        \includegraphics[width=0.23\textwidth]{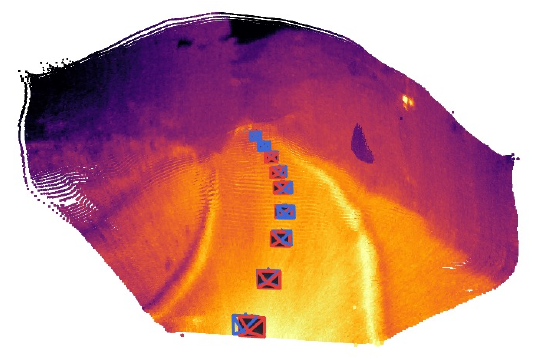} \\

        \includegraphics[width=0.23\textwidth]{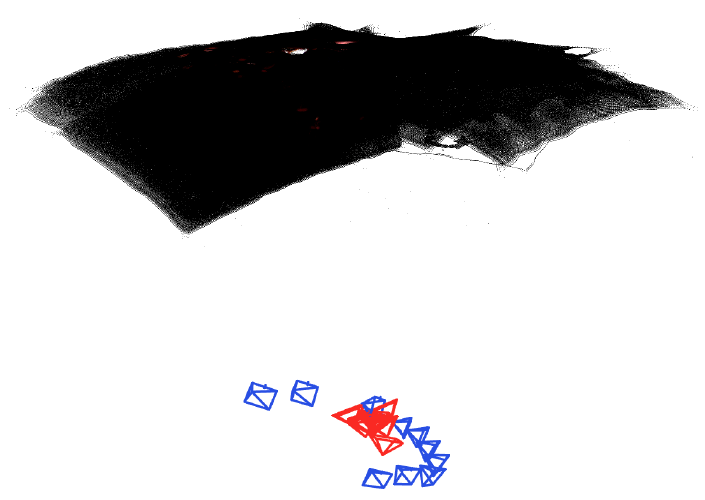} &
        \includegraphics[width=0.23\textwidth]{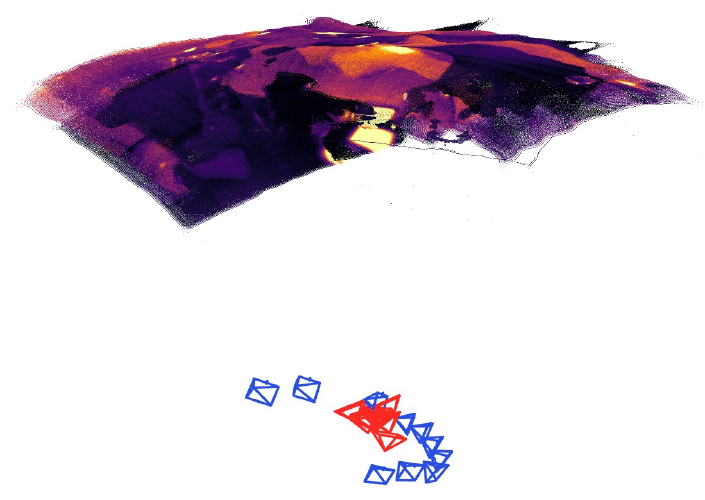} &
        \includegraphics[width=0.23\textwidth]{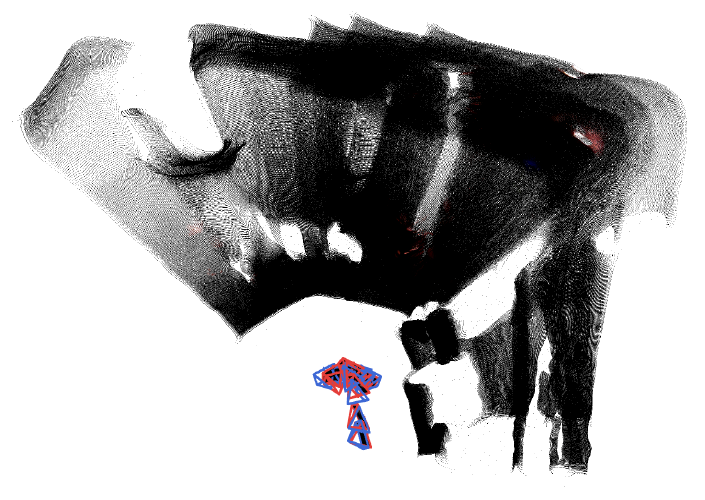} &
        \includegraphics[width=0.23\textwidth]{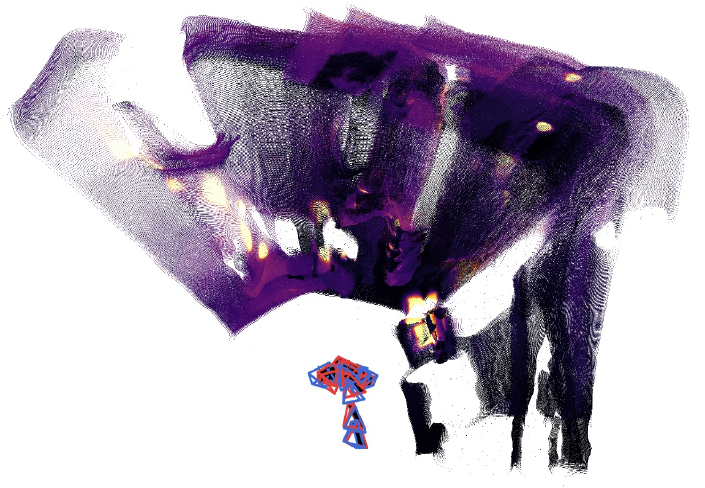} \\
    \end{tabular}

    \caption{Qualitative comparison between SEAR and our method. Blue and red cameras represent ground-truth and predicted camera poses, respectively.}
    \label{fig:additional_results}
\end{figure*}

Figure~\ref{fig:additional_results} compares additional dense 3D point-cloud reconstruction results between SEAR~\citep{skorokhodov2026sear}, an RGB-T feed-forward foundation model, and our proposed DarkVGGT on outdoor and indoor scenes.

In the nighttime outdoor sequences shown in the top two rows, both methods produce comparable point-map quality, likely because weak illumination cues such as streetlights remain available, allowing LoRA-based RGB-T adaptation to exploit residual RGB structure.
However, the camera-pose results show that DarkVGGT estimates trajectories closer to the ground truth than SEAR.

The gap becomes more pronounced in dark indoor scenes, where light and texture cues are nearly absent and effective fusion of thermal information into degraded RGB embeddings becomes crucial.
Under these conditions, SEAR struggles to recover reliable geometry.
In contrast, with the proposed thermal-to-RGB fusion modules, DarkVGGT robustly captures object-level depth variations and reconstructs plausible 3D geometry even in extremely dark indoor environments.

\subsection{Selected CTFF-only and full comparisons}
\label{app:ctff_full}

We examine selected sequence-level results to contextualize the aggregate comparison between LoRA+CTFF and the full model in Table~\ref{tab:ablation}. Adding GSTR reduces average AbsRel from 0.1422 to 0.1373 and increases RTA@30 from 0.86 to 0.89, while RRA@30 decreases from 1.00 to 0.99 and AUC@30 decreases from 68.9 to 66.3. Thus, the full model improves depth and translation accuracy at the reported threshold but does not uniformly improve pose performance.

On \texttt{indoor\_aggresive\_dark}, adding GSTR increases RTA@30 from 0.3733 to 0.4222 (13.1\% relative improvement), with the full model reaching AUC@30 of 19.3. On \texttt{indoor\_unstable\_dark}, AbsRel decreases from 0.1251 to 0.1228 and RTA@30 increases from 0.8259 to 0.8481 (2.7\% relative improvement). On STheReO, LoRA+CTFF and the full model achieve similar dataset-average AbsRel values of 0.1613 and 0.1600, respectively.

These selected results demonstrate local improvements alongside a lower aggregate AUC@30. Since AUC@30 integrates joint pose accuracy over thresholds, improvements in RTA@30 alone do not imply an improvement in AUC@30. This analysis is limited to the selected sequences and dataset average above.

\subsection{Controlled input and fusion baselines}
\label{app:controlled_baselines}

Table~\ref{tab:controlled_baselines} compares input modalities and fusion strategies on a separate controlled ViViD++ split. The test scenes are \texttt{indoor\_aggresive\_dark}, \texttt{indoor\_unstable\_dark}, \texttt{outdoor\_robust\_night2}, \texttt{indoor\_unstable\_global}, and \texttt{outdoor\_robust\_day2}. The last two scenes are moved from the main experiment's training split to evaluation. All fine-tuned rows use the same split, training schedule, and ground-truth supervision. DarkVGGT is retrained for this comparison. These results are not evaluations of the main experiment's checkpoint. ``Enh'' denotes CLAHE enhancement of RGB inputs. The naive multimodal baseline fine-tunes the AA blocks.

\begin{table}[!t]
\centering
\small
\caption{Controlled input and fusion comparison on a separate ViViD++ split. Fine-tuned methods share the split, schedule, and supervision. Dark and light columns report results on the three dark and two daylight test sequences, respectively. The top three values across all rows in each metric column are marked as \protect\colorbox{red!25}{\strut 1st}, \protect\colorbox{orange!25}{\strut 2nd}, and \protect\colorbox{yellow!35}{\strut 3rd}.}
\label{tab:controlled_baselines}
\setlength{\tabcolsep}{4pt}
\resizebox{0.8\linewidth}{!}{%
\begin{tabular}{llcccc}
\toprule
 & & \multicolumn{2}{c}{Dark} & \multicolumn{2}{c}{Light} \\
\cmidrule(lr){3-4}\cmidrule(lr){5-6}
Input / method & Setup & AbsRel$\downarrow$ & AUC@30$\uparrow$ & AbsRel$\downarrow$ & AUC@30$\uparrow$ \\
\midrule
RGB-only (raw) & Zero-shot & 0.2746 & 25.9 & 0.0977 & 71.6 \\
RGB-only + Enh & Zero-shot & 0.2841 & 24.9 & 0.0959 & 72.1 \\
RGB-only + Enh & Fine-tuned & 0.1736 & 28.8 & \secondcell{0.0729} & 73.0 \\
Thermal-only & Zero-shot & 0.2254 & \secondcell{57.4} & 0.2413 & 53.3 \\
Thermal-only & Fine-tuned & \bestcell{0.0940} & \bestcell{64.9} & 0.0914 & \secondcell{73.7} \\
Naive multimodal & Fine-tuned & \thirdcell{0.1722} & 29.1 & \bestcell{0.0718} & \thirdcell{73.6} \\
DarkVGGT (full) & Fine-tuned & \secondcell{0.1229} & \thirdcell{48.1} & \thirdcell{0.0732} & \bestcell{76.9} \\
\bottomrule
\end{tabular}%
}
\end{table}

Zero-shot RGB enhancement degrades performance on the dark split, increasing aggregate AbsRel from 0.2746 to 0.2841 and reducing AUC@30 from 25.9 to 24.9. Under matched fine-tuning, naive multimodal adaptation remains close to enhanced RGB-only (0.1722 versus 0.1736 AbsRel), whereas DarkVGGT reaches 0.1229 AbsRel and 48.1 AUC@30. Relative to naive multimodal adaptation, this is a 28.6\% reduction in dark AbsRel.

Thermal-only achieves better performance on the dark split. Its fine-tuned variant achieves 0.0940 AbsRel and 64.9 AUC@30, and its zero-shot variant also exceeds DarkVGGT in dark AUC@30 (57.4 versus 48.1). Thermal-only requires thermal input and cannot directly operate on RGB-only benchmarks such as ETH3D and ScanNet++. Where thermal input is available in daylight, fine-tuned thermal-only has approximately 25\% higher AbsRel than DarkVGGT (0.0914 versus 0.0732). DarkVGGT recovers approximately 64\% of the dark AbsRel gap between fine-tuned enhanced RGB-only and thermal-only, while remaining close to enhanced RGB-only in daylight AbsRel (0.0732 versus 0.0729) and achieving the highest daylight AUC@30 in this comparison (76.9). Naive multimodal adaptation achieves lower daylight AbsRel than DarkVGGT (0.0718 versus 0.0732). These results characterize a trade-off across illumination and sensor availability.

\subsection{Dark3R frame-sampling variance}
\label{app:dark3r_variability}

Dark3R provides a complementary evaluation using synthesized thermal inputs. The \texttt{chapel} and \texttt{kitchen} scenes are held out from training. Table~\ref{tab:dark3r_variability} reports per-sequence means and standard deviations over ten frame-sampling trials, supplementing the deterministic evaluation used in the main benchmark tables. These standard deviations characterize sensitivity to frame selection, not variability across independently trained models.

\begin{table}[!tb]
\centering
\small
\caption{Per-sequence performance on Dark3R dataset over 10 frame-sampling trials (mean $\pm$ standard deviation).}
\label{tab:dark3r_variability}
\setlength{\tabcolsep}{4pt}
\resizebox{\linewidth}{!}{%
\begin{tabular}{lcccccc}
\toprule
Scene & AUC@30$\uparrow$ & RRA@30$\uparrow$ & RTA@30$\uparrow$ & AbsRel$\downarrow$ & $\delta<1.25\uparrow$ & RMSE$_{\log}\downarrow$ \\
\midrule
\texttt{chapel} & $64.9\pm8.2$ & $1.000\pm0.000$ & $0.908\pm0.055$ & $0.114\pm0.013$ & $0.824\pm0.046$ & $0.163\pm0.016$ \\
\texttt{kitchen} & $46.7\pm3.5$ & $0.998\pm0.004$ & $0.959\pm0.022$ & $0.125\pm0.006$ & $0.868\pm0.020$ & $0.157\pm0.006$ \\
\bottomrule
\end{tabular}%
}
\end{table}

AUC@30 exhibits different sampling variability across the two scenes (standard deviations 8.2 and 3.5), whereas the reported AbsRel and RRA@30 standard deviations are at most 0.013 and 0.004, respectively. Pairwise pose evaluation aggregates $\binom{N}{2}$ frame pairs for $N$ sampled frames, so a difficult segment can affect multiple pairs simultaneously. This offers a possible explanation for sensitivity to frame selection, but the sampling experiment alone does not isolate its cause or establish training stability. These estimates are based on two held-out Dark3R scenes.

\subsection{Daylight scene performance on RGB-T datasets}
\label{app:rgbt_wellit_depth}

To verify that DarkVGGT's RGB-T fusion strategy remains effective under daylight conditions, we further evaluate depth estimation performance on three held-out well-lit sequences in Table~\ref{tab:depth_heldout_vivid}. We use the \texttt{city\_day1}, \texttt{city\_day2}, and \texttt{campus\_day2} sequences from ViViD++~\citep{lee2022vivid++} dataset. Since the extrinsics of these sequences are released as identity matrices, we additionally evaluate only depth estimation in this ablation study.

\begin{table*}[!tb]
\centering
\small
\caption{Quantitative comparison of depth estimation on well-lit RGB-T ViViD++ sequences. This shows that our thermal-to-RGB fusion method remains effective even in well-lit scenes.}
\label{tab:depth_heldout_vivid}
\setlength{\tabcolsep}{5pt}
\resizebox{\textwidth}{!}{%
\begin{tabular}{l *{3}{c} *{3}{c} *{3}{c}}
\toprule
 & \multicolumn{3}{c}{\texttt{city\_day1}}
 & \multicolumn{3}{c}{\texttt{city\_day2}}
 & \multicolumn{3}{c}{\texttt{campus\_day2}} \\
\cmidrule(lr){2-4}\cmidrule(lr){5-7}\cmidrule(lr){8-10}
Method
 & AbsRel$\downarrow$ & $\delta\!<\!1.25\uparrow$ & RMSE$_{\log}\!\downarrow$
 & AbsRel$\downarrow$ & $\delta\!<\!1.25\uparrow$ & RMSE$_{\log}\!\downarrow$
 & AbsRel$\downarrow$ & $\delta\!<\!1.25\uparrow$ & RMSE$_{\log}\!\downarrow$ \\
\midrule
MASt3R~\cite{leroy2024grounding}     &  \thirdcell{0.3133}    &  \thirdcell{0.5769}    &  \thirdcell{0.4079}    &  \thirdcell{0.2785}    &  \thirdcell{0.4284}    &  \thirdcell{0.3720}    &  \thirdcell{0.1613}    &  \secondcell{0.8154}    &  \thirdcell{0.2548}    \\
Dark3R~\cite{guo2026dark3r}          &  0.3479  &  0.4617    &  0.4703    &  0.4187    &  0.3587    &  0.5548    &  0.2268    &  0.6811    &  0.3157    \\
VGGT~\cite{wang2025vggt}             & \bestcell{0.1739} & \secondcell{0.7395} & \secondcell{0.2661} & \bestcell{0.1882} & \bestcell{0.7281} & \secondcell{0.3371} & \bestcell{0.1365} & \bestcell{0.8639} & \secondcell{0.2474} \\
DarkVGGT (ours)                      & \secondcell{0.1786} & \bestcell{0.7496} & \bestcell{0.2550} & \secondcell{0.1942} & \secondcell{0.7139} & \bestcell{0.3235} & \secondcell{0.1502} & \thirdcell{0.8153} & \bestcell{0.2328} \\
\bottomrule
\end{tabular}%
}
\end{table*}
DarkVGGT achieves the lowest $\mathrm{RMSE}_{\log}$ on all three held-out sequences.
Averaged over the three sequences, DarkVGGT reaches 0.1743 AbsRel, 0.7596 $\delta\!<\!1.25$, and 0.2704 $\mathrm{RMSE}_{\log}$, compared with VGGT's 0.1662, 0.7772, and 0.2835. 
DarkVGGT improves $\mathrm{RMSE}_{\log}$ at the cost of slightly worse average AbsRel and $\delta\!<\!1.25$, while largely preserving VGGT's RGB geometry performance.

\section{Experimental setup}

\subsection{RGB-T dataset processing}
\label{app:rgbt_processing}

Table~\ref{tab:dataset_stats} summarizes the number of samples used for our overall training and evaluation.

\begin{table}[!tb]
\centering
\small
\caption{Dataset split statistics used in the main experiments.}
\label{tab:dataset_stats}
\setlength{\tabcolsep}{6pt}
\resizebox{\linewidth}{!}{%
\begin{tabular}{lccccc}
\toprule
Dataset & Train Sequences & Train Frames & Eval Sequences & Eval Chunks & Eval Sampled Frames \\
\midrule
ViViD++~\citep{lee2022vivid++}  & 11 & 6{,}691 & 3 & 31 & 310 \\
STheReO~\citep{yun2022sthereo}  & 4  & 9{,}837 & 2 & 60 & 600 \\
Dark3R~\citep{guo2026dark3r}   & 6  & 2{,}260 & 2 & 2  & 50 \\
\midrule
Total    & 21 & 18{,}788 & 7 & 93 & 960 \\
\bottomrule
\end{tabular}
}
\end{table}

\paragraph{ViViD++.}
We train on 11 ViViD++ scenes and evaluate on three disjoint dark scenes.
The training split includes \texttt{indoor\_robust\_dark}, \texttt{outdoor\_robust\_night1}, \texttt{indoor\_aggresive\_global}, \texttt{indoor\_robust\_global}, \texttt{indoor\_unstable\_global}, \texttt{outdoor\_robust\_day1}, \texttt{outdoor\_robust\_day2}, \texttt{indoor\_robust\_varying}, \texttt{indoor\_aggresive\_local}, \texttt{indoor\_robust\_local}, and \texttt{indoor\_unstable\_local}, while the held-out set contains \texttt{indoor\_aggresive\_dark}, \texttt{indoor\_unstable\_dark}, and \texttt{outdoor\_robust\_night2}.
Sequence-mode training uses all frames from the selected scenes. We use refined depth maps \texttt{Depth\_RGB\_refined} and \texttt{Depth\_T\_refined} for training. Thermal images are percentile-clipped, min-max normalized, enhanced with CLAHE~\citep{reza2004realization}, and colorized with the Inferno colormap before being passed to the frozen DINOv2 encoder, while the grayscale thermal image is retained for edge-disagreement supervision. Both modalities are resized to a common resolution under the same crop/resize transform.

\paragraph{STheReO.}
We train on four STheReO scenes, \texttt{snu\_evening}, \texttt{kaist\_afternoon}, \texttt{snu\_afternoon}, and \texttt{valley\_afternoon}, comprising 44{,}612 raw frames in total, and evaluate on \texttt{kaist\_evening} and \texttt{valley\_evening} (14{,}586 raw frames). We use 9{,}837 training frames and 13{,}128 evaluation frames after trimming 5\% from each evaluation sequence end. Sequence-mode evaluation uses strict non-overlapping K=50 windows capped at 30 per scene, yielding 60 windows and 600 sampled frames. Raw STheReO sequences are converted into the canonical ViViD++ layout. For each RGB timestamp, we match the nearest thermal frame and LiDAR scan, interpolate the LiDAR-body pose to obtain camera-to-world poses, and undistort both modalities. Sparse depth maps are obtained by projecting the matched LiDAR scan into each camera, with optional dense refinements stored under \texttt{Depth\_RGB\_refined} and \texttt{Depth\_T\_refined}. During training, we disable static load-time subsampling and instead randomize the temporal spacing between nearby sampled frames with a stride drawn from $[2,4]$.

\begin{figure*}[!b]
    \centering
    \setlength{\tabcolsep}{2pt}
    \renewcommand{\arraystretch}{1.0}

    \begin{tabular}{@{}cccc@{}}
        {\scriptsize\text{Clean RGB}} &
        {\scriptsize\text{Dark RGB}} &
        {\scriptsize\text{Depth}} &
        {\scriptsize\text{Thermal}} \\[-0.2em]

        \includegraphics[width=0.235\textwidth]{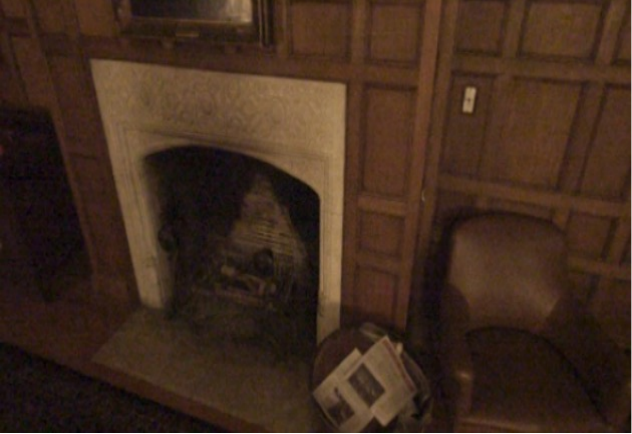} &
        \includegraphics[width=0.235\textwidth]{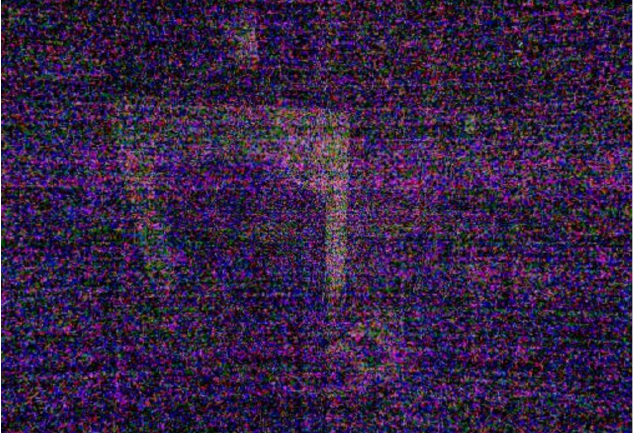} &
        \includegraphics[width=0.235\textwidth]{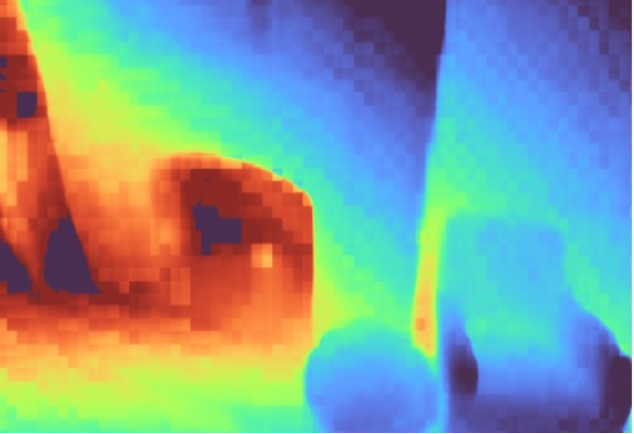} &
        \includegraphics[width=0.235\textwidth]{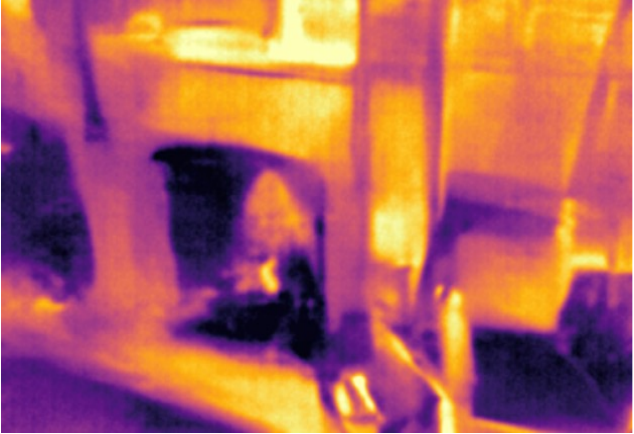} \\
    \end{tabular}
    \caption{Preprocessed Dark3R dataset training samples.}
    \label{fig:supp_dark3r_sample}
\end{figure*}

\paragraph{Dark3R.}
We train on the clean and dark variants of \texttt{university\_college}, \texttt{quad}, and \texttt{fireplace} (six sequences) and evaluate on \texttt{dark\_chapel} and \texttt{dark\_kitchen} under the public Dark3R protocol. Figure~\ref{fig:supp_dark3r_sample} shows our RGB-T conversion. As Dark3R lacks a calibrated dual-sensor rig, each exposure bracket yields a clean--dark pair with shared poses and released dense metric ground-truth depth. Clean RGB uses the cleanest sRGB image from the depth-annotation ISO folder, while Dark RGB uses the noisiest exposure (${\sim}40{\times}$ darker). ThermalGen~\citep{xiao2025thermalgen} generates paired pseudo-thermal supervision, and camera poses are converted from NeRF to OpenCV.

\subsection{Lingbot-Depth supervision and its scope}
\label{app:supervision_prior}

For ViViD++ and STheReO, Lingbot-Depth~\citep{tan2026masked} is used as an off-the-shelf completion tool to densify sparse LiDAR measurements into pixel-dense training targets and reduce optimization difficulties associated with sparse supervision. DarkVGGT and the matched fine-tuned SEAR baseline use the same refined supervision, whereas evaluation uses raw LiDAR ground truth rather than the refined targets. This parity applies to the matched training comparison, not to the off-the-shelf baselines.

Lingbot-Depth can introduce learned priors into the training targets. The contribution of depth refinement relative to training on raw sparse LiDAR, including any architecture-dependent effects, remains unquantified.

\subsection{Hyperparameters}
\label{app:hyperparam}

Table~\ref{tab:hyperparams} summarizes the key training hyperparameters used in DarkVGGT. We initialize the model from the public VGGT-1B checkpoint~\citep{wang2025vggt} and keep the backbone frozen throughout training. Multimodal adaptation is handled by the lightweight modules listed in Table~\ref{tab:hyperparams}, while the GSTR residual is initialized conservatively to avoid perturbing the pretrained RGB prior.

\begin{table}[!b]
\centering
\small
\caption{Key training hyperparameters for DarkVGGT.}
\label{tab:hyperparams}
\setlength{\tabcolsep}{6pt}
\begin{tabular}{ll}
\toprule
\multicolumn{2}{l}{\textit{Architecture / data}} \\
Backbone initialization & VGGT-1B, pretrained checkpoint \\
Image size / patch size & 518 / 14 \\
Clip lengths & $\{2,3,4,6,12\}$ \\
Max images per GPU & 24 \\
\midrule
\multicolumn{2}{l}{\textit{Optimization}} \\
Optimizer & AdamW \\
Learning rate & $5\times10^{-5}$ \\
Weight decay & $10^{-2}$ \\
Schedule & 10\% linear warmup, 90\% cosine decay \\
Mixed precision & bfloat16 \\
Gradient accumulation & 2 \\
Epochs / steps & 10 / 2000 \\
Effective steps (accum.\ $\times$ steps) & $4000 = 2 \times2000$ \\
\midrule
\multicolumn{2}{l}{\textit{Adaptation modules}} \\
LoRA rank / alpha & 64 / 128 \\
Geometry-shared ranks & $(d_s,d_p,d_{mi})=(96,192,128)$ \\
GSTR active blocks & $k=\mathbf{8}$ \\
GSTR residual init $\alpha_{\mathrm{gstr}}$ & $\sigma(-2.2)\approx 0.10$ \\
\midrule
\multicolumn{2}{l}{\textit{Loss weights}} \\
Camera $\mathcal{L}_{\mathrm{cam}}$ & 5.0 \\
Depth $\mathcal{L}_{\mathrm{depth}}$ (log-norm) & 1.0 \\
Residual allocation $\mathcal{L}_{\mathrm{sparse}}$ & 0.03 \\
Edge disagreement $\mathcal{L}_{\mathrm{edge}}$ & 0.01 \\
Orthogonality $\mathcal{L}_{\mathrm{ortho}}$ & 0.003 \\
RGB-prior preservation $\mathcal{L}_{\mathrm{drop}}$ & 0.5 \\
GSTR recon $\mathcal{L}_{\mathrm{recon}}$ & 0.01 $\rightarrow$ 0.10 (cosine, step $0$–$1500$ of $4000$) \\
GSTR distill $\mathcal{L}_{\mathrm{distill}}$ & 0.01 $\rightarrow$ 0.10 (cosine, step $0$–$1500$ of $4000$) \\
GSTR decorrelation $\mathcal{L}_{\mathrm{decorr}}$ & 0 $\rightarrow$ $2\!\times\!10^{-5}$ (cosine, step $1000$–$3000$ of $4000$) \\
\midrule
\multicolumn{2}{l}{\textit{Multimodal training}} \\
Thermal dropout & $p=0.5$, RGB-dominant batches only \\
\bottomrule
\end{tabular}
\end{table}

The auxiliary losses are kept small relative to the main VGGT supervision. The CTFF regularizers remain weak, the GSTR reconstruction and distillation terms use cosine warmup, and the decorrelation term is delayed and ramped gradually. In the final configuration, we omit the explicit patch-level correction loss and apply RGB-prior preservation only on thermal-dropout steps. Thermal dropout is restricted to batches whose clips all come from a predefined set of daylight training sequences, identified by scene-name matching, yielding an effective dropout rate of about 25\%. At inference, no scene-name rules are used. The learned feature-conditioned gates modulate thermal injection.

\subsection{Evaluation protocol}
All main benchmark results are reported in sequence mode on held-out scenes only. For each held-out scene, we form strict non-overlapping $K{=}50$-frame windows and uniformly sample $N{=}10$ frames per window, capped at 30 windows per scene. ViViD++ scenes are short and never reach this cap (up to 20 windows per scene), whereas STheReO sequences are long driving recordings for which the cap is active, yielding 30 windows per scene. For the Dark3R dataset, we use a single deterministic chunk per scene, sampling the full scene with a stride of $20$ to obtain approximately $25$ frames. Depth metrics are averaged over the sampled evaluation frames, while pose metrics are pooled over all frame pairs within each sampled window.

\end{document}